\documentclass{llncs}
\usepackage{geometry}
\usepackage{ucs}
\usepackage[utf8x]{inputenc}
\usepackage[T1]{fontenc}
\usepackage{url}
\urldef\myurl\url{http://pl.wikipedia.org/wiki/Lech_Wa%C5%82%C4%99sa}
\usepackage{multirow}
\usepackage{amsmath}
\usepackage{lscape}
\usepackage{graphicx}
\let\chapter\section

\usepackage[boxed,vlined]{algorithm2e}

\usepackage[]{natbib}

\begin{document}
\title{Boosting Question Answering by Deep Entity Recognition}

\titlerunning{Gathering Knowledge for Question Answering Beyond Named Entities} 

\author{Piotr Przybyła\footnote{The study was supported by research fellowship within "Information technologies: research and their interdisciplinary applications" agreement number POKL. 04.01.01-00-051/10-00.}}
\authorrunning{Piotr Przybyła}   
%
\tocauthor{Piotr Przybyła (Institute of Computer Science, Polish Academy of Sciences)}
\institute{Institute of Computer Science, Polish Academy of Sciences,\\
Warsaw, Poland,\\
\email{P.Przybyla@phd.ipipan.waw.pl}}

\maketitle              

\begin{abstract}
In this paper an open-domain factoid question answering system for Polish, RAFAEL, is presented. The system goes beyond finding an answering sentence; it also extracts a single string, corresponding to the required entity. Herein the focus is placed on different approaches to entity recognition, essential for retrieving information matching question constraints. Apart from traditional approach, including named entity recognition (NER) solutions, a novel technique, called Deep Entity Recognition (DeepER), is introduced and implemented. It allows a comprehensive search of all forms of entity references matching a given WordNet synset (e.g. an impressionist), based on a previously assembled entity library. It has been created by analysing the first sentences of encyclopaedia entries and disambiguation and redirect pages. DeepER also provides automatic evaluation, which makes possible numerous experiments, including over a thousand questions from a quiz TV show answered on the grounds of Polish Wikipedia. The final results of a manual evaluation on a separate question set show that the strength of DeepER approach lies in its ability to answer questions that demand answers beyond the traditional categories of named entities.  
\end{abstract}

\section{Introduction}

A Question Answering (QA) system is a computer program capable of understanding questions in a natural language, finding answers to them in a knowledge base and providing answers in the same language. So broadly defined task seems very hard; \citet{Shapiro1992} describes it as {\em AI-Complete}, i.e. equivalent to building a general artificial intelligence. Nonetheless, the field has attracted a lot of attention in Natural Language Processing (NLP) community as it provides a way to employ numerous NLP tools in an exploitable end-user system. It has resulted in valuable contributions within TREC competitions \citep{Dang2008} and, quite recently, in a system called \textit{IBM Watson} \citep{Ferrucci2010}, successfully competing with humans in the task.

However, the problem remains far from solved. Firstly, solutions designed for English are not always easily transferable to other languages with more complex syntax rules and less resources available, such as Slavonic. Secondly, vast complexity and formidable hardware requirements of \textit{IBM Watson} suggest that there is still a room for improvements, making QA systems smaller and smarter.

This work attempts to contribute in both of the above areas. It introduces RAFAEL ({\em RApid Factoid Answer Extraction aLgorithm}), a complete QA system for Polish language. It is the first QA system designed to use an open-domain plain-text knowledge base in Polish to address factoid questions not only by providing the most relevant sentence, but also an entity, representing the answer itself. The Polish language, as other Slavonic, features complex inflection and relatively free word order, which poses additional challenges in QA. Chapter \ref{RAFAEL} contains a detailed description of the system architecture and its constituents.

In the majority of such systems, designers' attention focus on different aspects of a sentence selection procedure. Herein, a different idea is incorporated, concentrating on an entity picking procedure. It allows to compare fewer sentences, likely to contain an answer. To do that, classical Named Entity Recognition (NER) gets replaced by Deep Entity Recognition. {\em DeepER}, introduced in this work, is a generalisation of NER which, instead of assigning each entity to one of several predefined NE categories, assigns it to a WordNet synset.

For example, let us consider a question: \textit{Which exiled European monarch returned to his country as a prime minister of a republic?}. In the classical approach, we recognise the question as concerning a person and treat all persons found in texts as potential answers. Using DeepER, it is possible to limit the search to persons being monarchs, which results in more accurate answers. In particular, we could utilise information that \textit{Simeon II} (our answer) is a tsar; thanks to WordNet relations we know that it implies being a monarch. DeepER is a generalisation of NER also from another point of view -- it goes beyond the classical named entity categories and treats all entities equally. For example, we could answer a question \textit{Which bird migrates from the Arctic to the Antarctic and back every year?}, although \textit{arctic tern} is not recognized as NE by NER systems. Using DeepER, we may mark it as a seabird (hence a bird) and include among possible answers. Chapter \ref{DeepER} outlines this approach.

The entity recognition process requires an {\em entities library}, containing known entities, their text representations (different ways of textual notation) and WordNet synsets, to which they belong. To obtain this information, the program analyses definitions of entries found in encyclopaedia (in this case the Polish Wikipedia). In previous example, it would use a Wikipedia definition: \textit{The Arctic Tern (Sterna paradisaea) is a seabird of the tern family Sternidae.}  This process, involving also redirect and disambiguation pages, is described in section \ref{Library}. Next, having all the entities and their names, it suffices to locate their mentions in a text. The task (section \ref{Recognition}) is far from trivial because of a complicated named entity inflection in Polish (typical for Slavonic languages, see \citep{Przepiorkowski2007}).

DeepER framework provides also another useful service, i.e. automatic evaluation. Usually QA systems are evaluated by verifying accordance between obtained and actual answer based on a human judgement. Plain string-to-string equality is not enough, as many entities have different text representations, e.g. \textit{John F. Kennedy} is as good as \textit{John Fitzgerald Kennedy} and \textit{John Kennedy}, or \textit{JFK} (again, the nominal inflection in Polish complicates the problem even more). However, with DeepER, a candidate answer can undergo the same recognition process and be compared to the actual expected {\em entity}, not {\em string}.

Thanks to automatic evaluation vast experiments requiring numerous evaluations may be performed swiftly; saving massive amount of time and human resources. As a test set, authentic questions from a popular Polish quiz TV show\footnote{\textit{Jeden z dziesięciu}} are used. Results of experiments, testing (among others) the optimal context length, a number of retrieved documents, a type of entity recognition solution, appear in section \ref{experiments}.

To avoid overfitting, the final system evaluation is executed on a separate test set, previously unused in development, and is checked manually. The results are shown in section \ref{final} and discussed in chapter \ref{discussion}. Finally, chapter \ref{conclusions} concludes the paper.

\section{RAFAEL}
\label{RAFAEL}
As stated in previous chapter, RAFAEL is a computer system solving a task of Polish text-based, open-domain, factoid question answering. It means that provided questions, knowledge base and returned answers are expressed in Polish and may belong to any domain. The system analyses the knowledge base, consisting of a set of plain text documents, and returns answers (as concise as possible, e.g. a person name), supplied with information about supporting sentences and documents.

What are the kinds of requests that fall into the category of {\em factoid questions}? For the purpose of this study, it is understood to include the following types:
\begin{itemize}
\item \textbf{Verification} questions, which could be answered by providing a single boolean value (true or false), e.g. \textit{Did Lee Oswald kill John F. Kennedy?},
\item \textbf{Option} questions, requiring to select one of available options, e.g. \textit{Which one killed John F. Kennedy: Lance Oswald or Lee Oswald?},
\item \textbf{Named entity} questions, expected to be answered by a single named entity, e.g. \textit{When was John F. Kennedy killed?},
\item \textbf{Unnamed entity} questions, similar to above, but the expected entity does not belong to traditional named entity types, e.g. \textit{What did Lee Oswald use to kill John F. Kennedy?},
\item \textbf{Other name} questions, asking for another name of a given named entity, e.g. \textit{What nickname did John F. Kennedy use during his military service?},
\item \textbf{Multiple named entities} questions, to be answered by a list of named entities, e.g. \textit{Which U.S. presidents were assassinated in office?}. 
\end{itemize} 
Although the above list rules out many challenging types of questions, demanding more elaborate answers (e.g. \textit{Why was JFK killed?}, \textit{What is a global warming?}, \textit{How to build a fence?}), it still involves very distinct problems. Although RAFAEL can recognize factoid questions from any of these types and find documents relevant to them (see more in section \ref{analysis} and \citep{Przybyla}), its answering capabilities are limited to those requesting single \textbf{unnamed entities} and \textbf{named entities}. In this document, they are called \textbf{entity questions}.

The task description here is similar to the TREC competitions and, completed with test data described in section \ref{data}, could play a similar role for Polish QA, i.e. provide a possibility to compare different solutions of the same problem. More information about the task, including its motivation, difficulties and a feasibility study for Polish could be found in \citep{Przybya2012}.

\subsection{Related work}
The problem of Question Answering is not new to the Polish NLP community (nor working on other morphologically rich languages), but none of studies presented so far coincides with the notion of plain text-based QA presented above.

First Polish QA attempts date back to 1985, when \citet{Vetulani1988} presented a Polish interface to ORBIS database, containing information about the solar system. The database consisted of a set of PROLOG rules and the role of the system (called POLINT) was to translate Polish questions to appropriate queries. Another early solution, presented by \citet{Duclaye2002}, could only work in a restricted domain (business information).

A system dealing with a subset of the  TREC tasks was created for Bulgarian by \citet{Tanev2004}. His solution answers only three types of questions: Definition, Where-Is and Temporal. He was able to achieve good results with 100 translated TREC questions, using several manually created answer patterns, without NER or any semantic information. Another system for Bulgarian \citep{Simov2005} participated in the CLEF 2005 competition. Its answer extraction module bases on partial grammars, playing a role of patterns for different types of questions. They could answer correctly 37 of 200 questions, of which only 16 belong to the factoid type. Previously the same team \citep{Osenova2004} took part in a Bulgarian-English track of the CLEF 2004, in which Bulgarian questions were answered using English texts.

A QA solution was also created for Slovene \citep{Ceh2009}. The task there is to answer students' questions using databases, spreadsheet files and a web service. Therefore, it differs from the problem discussed above by limited domain (issues related to a particular faculty) and the non-textual knowledge base. Unfortunately, no quantitative results are provided in this work.

More recently, several elements of a Polish QA system called {\em Hipisek} were presented by \citet{Walas2010}. It bases on a fairly common scheme of transforming a question into a search query and finding the most appropriate sentence, satisfying question constrains. Unfortunately, a very small evaluation set (65 question) and an unspecified knowledge base (gathered by a web crawler) make it difficult to compare the results. In their later works \citep{Walas2012,Walas2011}, the team concentrated on spatial reasoning using a knowledge base encoded as a set of predicates.

The approach presented by \citet{Piechocinski2005} is the closest to the scope of this work, as it includes analysis of Polish Wikipedia content and evaluation is based on questions translated from a TREC competition. Unfortunately, it heavily relies on a structure of Wikipedia entries, making it impossible to use with an arbitrary textual corpus.

A non-standard approach to answer patterns has been proposed by \citet{Konopik2010}. In their Czech open-domain QA system they used a set of templates associated with question types, but also presented a method to learn them semi-automatically from search results. \citet{Peshterliev2011} in their Bulgarian QA system concentrated on semantic matching between between a question and a possible answer checked using dependency parsing. However, they provide no data regarding an answering precision of the whole system.

The last Polish system worth mentioning has been created by \citet{Radziszewski2013}. Generally, their task, called Open Domain Question Answering (ODQA), resembles what is treated here, but with one major difference. A document is considered an answer; therefore they focus on improving ranking in a document retrieval stage. They have found out that it could benefit from taking nearness of query terms occurrences into account. 

As some of Slavonic languages lack necessary linguistic tools and resources, only partial solutions of QA problems exist for them, e.g. document retrieval for Macedonian \citep{Armenska2010}, question classification for Croatian \citep{Lombarovic2011} or answer validation for Russian \citep{Solovyev2013}.

\subsection{System Architecture}
\begin{figure}
  \centering
    \includegraphics[width=13.5cm]{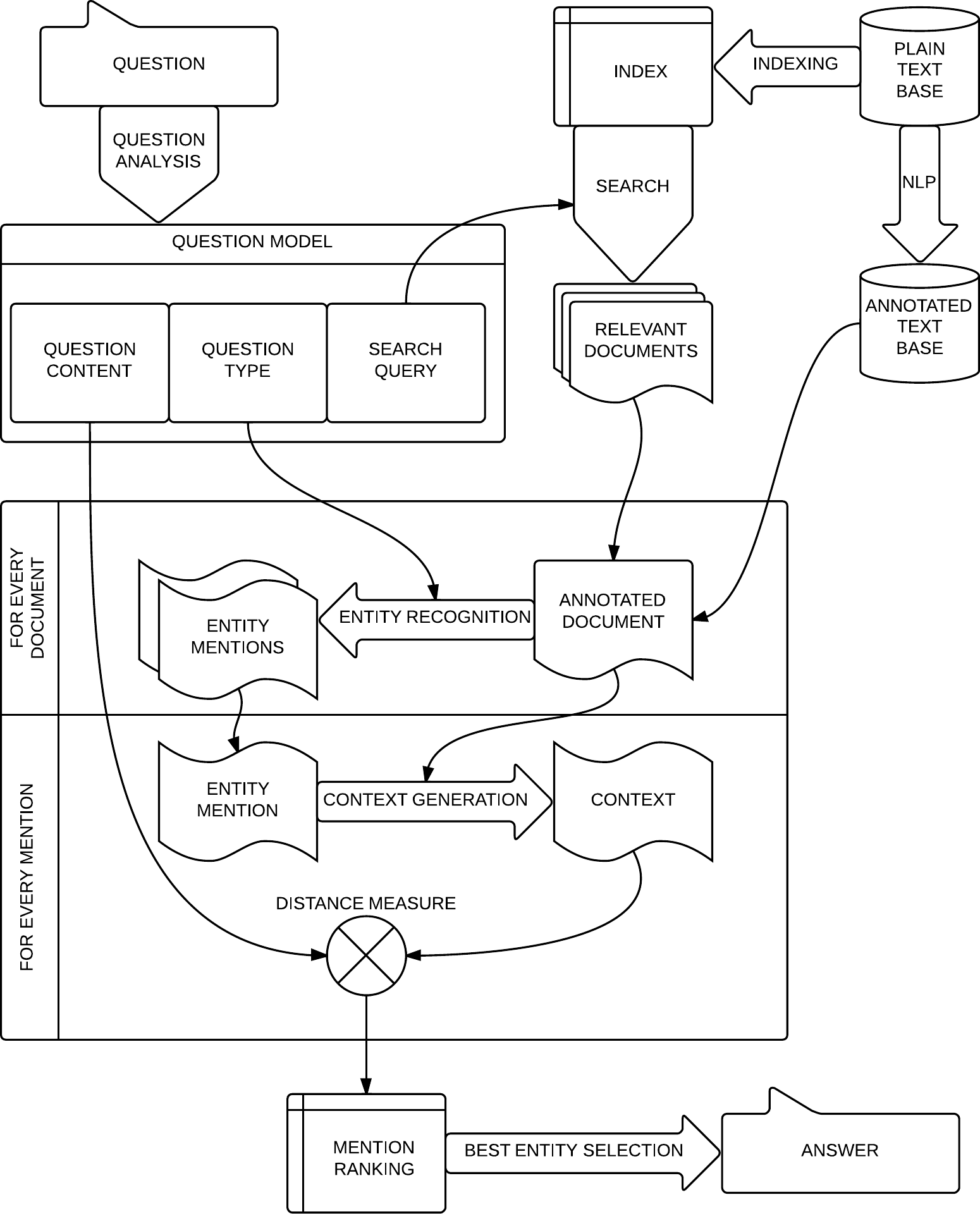}
  \caption{Overall architecture of the QA system -- RAFAEL. See descriptions of elements in text.}
  \label{fig:arch}  
\end{figure}

A general architectural scheme of RAFAEL (figure \ref{fig:arch}) has been inspired by similar systems developed for English; for examples see works by \citet{Hovy2000} and \citet{Moldovan2000}.

Two of the steps in the diagram concern offline processing of a knowledge base. Firstly, it is indexed by a search engine to ensure efficient searching in further stages (INDEXING). Secondly, it may be annotated using a set of tools (NLP), but this could also happen at an answering stage for selected documents only.

After the system receives a question, it gets analysed (QUESTION ANALYSIS) and transformed into a data structure, called \textit{question model}. One of its constituents, a search query, is used to find a set of documents, which are probably appropriate for the current problem (SEARCH). For each of the documents, all entity mentions compatible with an obtained question type (e.g. monarchs), are extracted (ENTITY RECOGNITION). For each of them, a context is generated (CONTEXT GENERATION). Finally, a distance between a question content and the entity context is computed to asses its relevance (DISTANCE MEASURE). All the mentions and their distance scores are stored and, after no more documents are left, used to select the best match (BEST ENTITY SELECTION). The system returns the entity, supplied with information about a supporting sentence and a document, as an answer.

\subsection{Knowledge Base Processing}
\label{preprocessing}

Knowledge base (KB) processing consists of two elements: indexing and annotating. The objective of the first is to create an index for efficient searching using a search engine. In the system, {\em Lucene} 3.6\footnote{Available from \url{http://lucene.apache.org/}.} is used to build two separate full-text indices: regular and stemmed using a built-in stemmer for Polish, {\em Stempel} \citep{Galambos2001}.

Secondly, texts go through a cascade of annotation tools, enriching it with the following information:
\begin{itemize}
\item Morphosyntactic interpretations (sets of tags), using {\em Morfeusz} 0.82 \citep{Wolinski2006a},
\item Tagging (selection of the most probable interpretation), using a transformation-based learning tagger, {\em PANTERA} 0.9.1 \citep{Acedanski2010},
\item Syntactic groups (possibly nested) with syntactic and semantic heads, using a rule-based shallow parser {\em Spejd} 1.3.7 \citep{Przepiorkowski2008} with a Polish grammar, including improved version of modifications by \citet{Degorski2012}, enabling lemmatisation of nominal syntactic groups,
\item Named entities, using two available tools: {\em NERF} 0.1 \citep{Savary2012} and {\em Liner2} 2.3 \citep{Marcinczuk2012}.
\end{itemize}
All the annotations are stored in a variant of {\em TEI P5} standard, designed for the National Corpus of Polish \citep{Przepiorkowski2012}. As noted previously, the process of annotating is not indispensable at the stage of offline KB processing; it could be as well executed only on documents returned from the search engine (for example see \textit{Webclopedia} by \citet{Hovy2000} or \textit{LASSO} by \citet{Moldovan2000}). However, since during evaluation experiments the same documents undergo the process hundreds of times, it seems reasonable to process the whole KB only once.

\subsection{Question Analysis}
\label{analysis}

The goal of question analysis is to examine a question and extract  all the information that suffices for answer finding. A resulting data structure, called {\em question model}, contains the following elements:
\begin{enumerate}
\item {\em Question type} -- a description of expected answer type, instructing the system, what type of data could be returned as an answer. It has three levels of specificity:
\begin{enumerate}
\item {\em General question type} -- one of the types of factoid questions, enumerated at the beginning of this chapter,
\item {\em Named entity type} -- applicable only in case general type equals {\em named entity}. Possible values are the following: place, continent, river, lake, mountain, mountain range, island, archipelago, sea, celestial body, country, state, city, nationality, person, first name, last name, band, dynasty, organisation, company, event, date, century, year, period, number, quantity, vehicle, animal, title.
\item {\em Focus synset} -- applicable in case of entity questions; a WordNet synset, to which a question focus belongs; necessary for DeepER. 
\end{enumerate}
\item {\em Search query} -- used to find possibly relevant documents,
\item {\em Question content} -- the words from question which are supposed to appear also in context of an answer.
\end{enumerate}

The task presented above, called {\em question classification}, is an example of text classification with very short texts. It could be tackled by a general-purpose classifier; for example, \citet{Ceh2009} used SVMs (Support Vector Machines) for closed-domain Slovene QA system; \citet{Li2002} employed SNoW (Sparse Network of Winnows) for hierarchical classification of TREC questions. For Polish results are not satisfactory \citep{Przybyla} because of data sparsity.

However, sometimes a solution seems quite evident, as part of the question types enforce its structure. For example, when it begins with {\em Who} or {\em When}, it belongs to {\em person} and {\em date} question types, respectively. That is why a set of 176 regular expressions (in case of RAFAEL) suffices to deal with them. They match only a subset of questions (36.15 per cent of the training set), but are highly unambiguous (precision of classification equals 95.37 per cent). Nevertheless, some \citep{Lee2005} use solely such patterns, but need a great number of them (1,273).

\begin{figure}
  \centering
    \includegraphics[width=13.5cm]{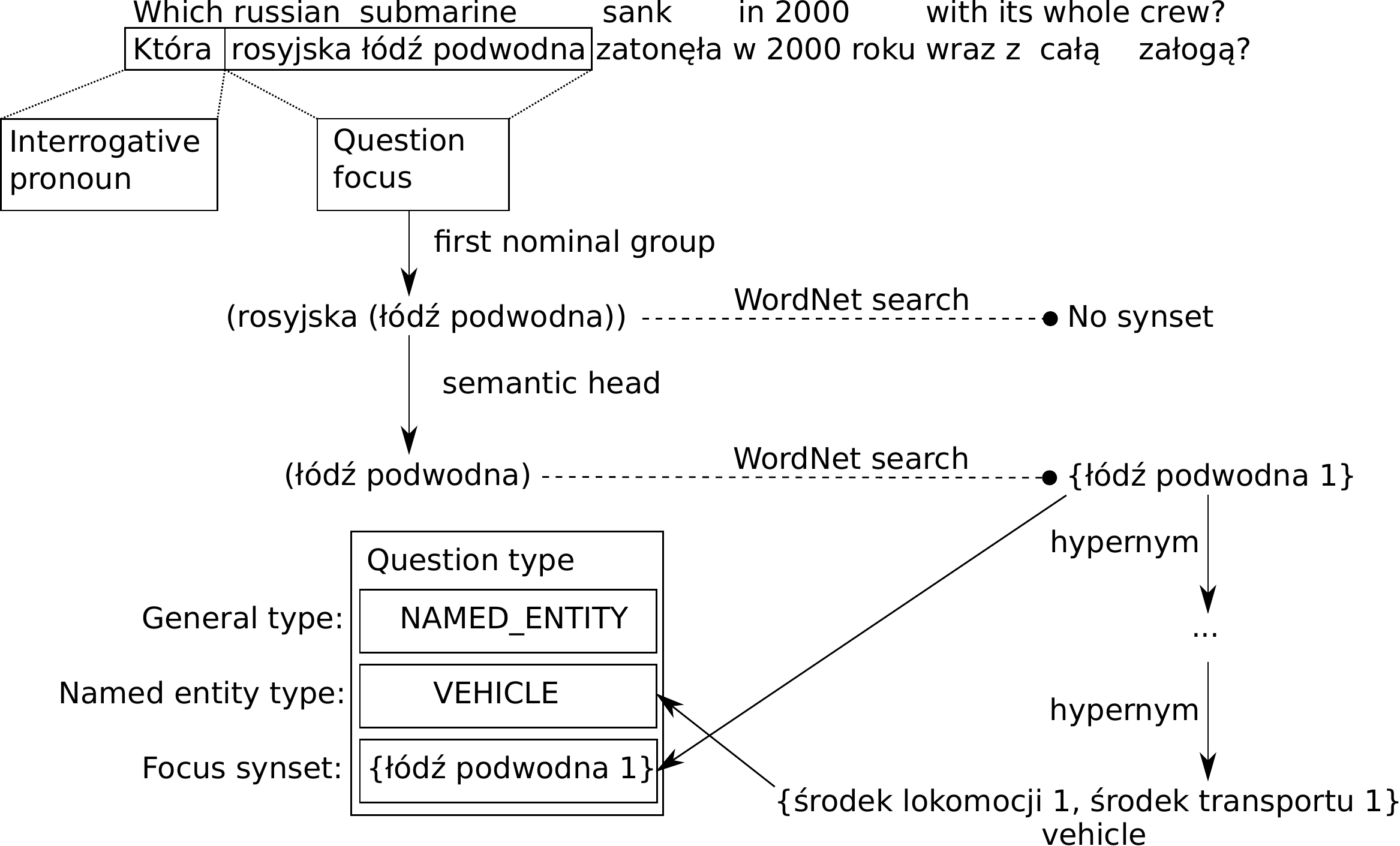}
  \caption{Outline of a question focus analysis procedure used to determine an entity type in case of ambiguous interrogative pronouns.}
  \label{fig:focus}  
\end{figure}
Unfortunately, most of entity questions are ambiguous, i.e. it is not enough to inspect an interrogative pronoun to find an answer type. They may begin with {\em what} or {\em which}, followed by a {\em question focus}. For example, let us consider a question \textit{Which russian submarine sank in 2000 with its whole crew?}. Its focus (\textit{russian submarine}) carries information that the question could be answered by a named entity of type {\em vehicle}. The whole process of focus analysis is shown in figure \ref{fig:focus}. The first nominal group after a pronoun\footnote{RAFAEL includes a manually created list of \textit{opening constructions} to be ignored if appearing right after a pronoun, such as \textit{typ} (type of) or \textit{spośród} (out of).} serves as a possible lexeme name in {\em plWordNet} 2.1 \citep{Maziarz2012}. As long as there are no results, it gets replaced by its semantic head. When a matching lexeme exists in WordNet, a set of all its hypernyms is extracted. If any of the elements in the set correspond to one of the named entity types, this type is recorded in the question model. Otherwise the general question type takes the value {\em unnamed entity}. A WordNet-assisted focus analysis was also implemented in one of solutions participating in a TREC competition \citep{Harabagiu2001}.

{\em Search query} generation is described in the next chapter. The last element of a question model, called {\em question content}, contains segments, which are to be compared with texts to find the best answer. It includes all the words of the interrogative sentence except those included in the matched pattern ({\em Which}, {\em ?}) and the focus ({\em submarine}). In our example the following are left: {\em russian}, {\em sank}, {\em in}, {\em 2000}, {\em with}, {\em its}, {\em whole}, {\em crew}. An entity mention, which context resembles this set, will be selected as an answer (see details in section \ref{selection}).

The question analysis stage explained above follows a design presented in previous works \citep{Przybyla,Przybya2013}, where more details could be found. The major difference lies in result processing -- an original synset is not only projected to one of the named entity types, but also recorded as a {\em focus synset} in question type, utilised in DeepER to match entity types. In our example, it would only consider submarines as candidate answers.

\subsection{Document Retrieval}

The use of search engines in QA systems is motivated mainly by performance reasons. Theoretically, we could analyse every document in a text base and find the most relevant to our query. However, it would take excessive amount of time to process the documents, majority of which belong to irrelevant domains (839,269 articles in the test set). A search engine is used to speed up the process by selecting a set of documents and limiting any further analysis to them.

As described in section \ref{preprocessing}, a knowledge base is indexed by {\em Lucene} offline. Given a question, we need to create a search query. The problem is that an answer in the knowledge base is probably expressed differently than the question. Hence, a query created directly from words of the question would not yield results, unless using a highly-redundant KB, such as the WWW (for this type of solution see \citep{Brill2002a}). Therefore, some of the query terms should be dropped -- based on their low IDF \citep{Katz2003} or more complex heuristics \citep{Moldovan2000}. On the other hand, the query may be expanded with synonyms \citep{Hovy2000} or derived morphological forms \citep{Katz2003}.

Finally, we need to address term matching issue -- how to compare a query keyword and a text word in a morphologically-rich language, such as Polish? Apart from exact match, it also is possible to use a stemmer or fuzzy queries, available in \textit{Lucene} (accepting a predefined Levenshtein distance between matching strings).

Previous experiments \citep{Przybya2013} led to the following query generation procedure:
\begin{enumerate}
\item Remove all words matched by a regular expression at the classification stage ({\em What}, {\em Which}, etc.),
\item Keep a question focus,
\item Connect all the remaining words by OR operator,
\item Use fuzzy term matching strategy with absolute distance equal 3 characters and fixed prefix.
\end{enumerate}

{\em Lucene} handles a query and yields a ranked document list, of which N first get transferred to further analysis. The influence of value of N on answering performance is evaluated in section \ref{experiments}.

\subsection{Entity Recognition}
\label{er}

Having a set of proposed documents and a question type, the next step is to scan them and find all mentions of entities with appropriate types. RAFAEL includes two approaches to the problem: classical Named Entity Recognition (NER) and novel Deep Entity Recognition. 

Three NERs for Polish are employed: {\em NERF}, {\em Liner2} and {\em Quant}. {\em NERF} \citep{Savary2012} is a tool designed within the project of the National Corpus of Polish and bases on linear-chain conditional random fields (CRF). It recognizes 13 types of NEs, possibly nested (e.g. \textit{Warsaw} in \textit{University of Warsaw}). {\em Liner2} \citep{Marcinczuk2012} also employs CRFs, but differentiates NEs of 56 types (which could be reduced to 5 for higher precision). Annotation using both of the tools happens offline within the KB preprocessing, so in the currently described stage it suffices to browse the annotations and find matching entities. As the above tools lack recognition of quantitative expressions, a new one has been developed especially for RAFAEL and called {\em Quant}. It is able to handle both numbers and quantities (using \textit{WordNet}) in a variety of notations.

Appendix A contains details of implementation of named entity recognition in RAFAEL, including a description of {\em Quant} and a mapping between question types and named entity types available in {\em NERF} and {\em Liner2}. An alternative being in focus of this work, i.e. DeepER approach, is thorougly discussed in chapter \ref{DeepER}. 

RAFAEL may use any of the two approaches to entity recognition: NER (via {\em NERF}, {\em Liner2} and {\em Quant}) or novel DeepER; this choice affects its overall performance. Experiments showing precision and recall of the whole system with respect to applied entity recognition technique are demonstrated in section \ref{experiments}. 

\subsection{Mention selection}
\label{selection}

When a list of entity mentions in a given document is available, we need to decide which of them most likely answers the question. The obvious way to do that is to compare surroundings of every mention with the content of the question. The procedure consists of two steps: context generation and similarity measurement.

\subsubsection{Context generation}
\label{context}

The aim of a context generation step is to create a set of segments surrounding an entity, to which they are assigned. Without capabilities of full text understanding, two approximate approaches seem legitimate:
\begin{itemize}
\item Sentence-based -- for a given entity mention, a sentence in which it appears, serves as a context,
\item Segment-based -- for a given entity mention, every segment sequence of length {\em M}, containing the entity, is a context.
\end{itemize}
Both of them have some advantages: relying on a single sentence ensures relation between an entity and a context, whereas the latter provides possibility of modifying context length. Obviously, the value of {\em M} should be proportional to question (precisely, its content) length.

The method of treating sentences as a context has gained most popularity (see work of \citet{Yih2013}), but a window of fixed size also appears in the literature; for example \citet{Katz2003} used one with M=140 bytes.

The context generation is also related to another issue, i.e. anaphoric expressions. Some  segments (e.g. {\em this}, {\em him}, {\em they}) may refer to entities that occurred earlier in a text and therefore harm a similarity estimation. It could be tackled by applying anaphora resolution, but a solution for Polish \citep{Kopec2012} remains in an early stage. Observations show that the majority of anaphora refer to an entity in a document title, so the problem is partially bypassed by adding a title to a context.

An influence of the context generation techniques on final results is shown in section \ref{experiments}. 

\subsubsection{Similarity measurement}

To measure a similarity between a question content (explained in section \ref{analysis}) and an entity context (generated by the procedures in previous section), a Jaccard similarity index \citep{Jaccard1901} is computed. However, not all word co-occurrences matter equally (e.g. compare {\em this} and {\em Honolulu}), so word weights are used:
\[
\text{Sim}_w(A,B)=\frac{\sum_{i\in A\cap B}w_i}{\sum_{i\in A\cup B}w_i}
\]
The sets $A$ and $B$ contain segments in base forms, whereas $w_i$ denotes a weight of an $i$-th base form, equal to its scaled IDF computed on a document set $D$: 
\[
w_i=\frac{\log \frac{|D|}{|\{d:i \in d\}|}}{\max\limits_i \log \frac{|D|}{|\{d:i \in d\}|}}
\]
The Jaccard index is a popular solution for sentence similarity measurement in QA (for example see a system by \citet{Ahn2004}). In case of selecting relevant documents, cosine measure is also applied. \citet{Radziszewski2013} compared it to Minimal Span Weighting (MSW) and observed that the latter performs better, as it takes into account a distance between matched words. A study of different techniques for sentence similarity assessment could be found in \citep{Yih2013}.

\subsubsection{Answer generation}
\label{generation}

At this stage, a large set of pairs of entity mention and its contexts with scores assigned, is available. Which of them answers the question? Choosing the one with the highest score seems an obvious solution, but we could also aggregate scores of different mentions corresponding to the same answer (entity), e.g. compute their sum or mean. However, such experiments did not yield improvement, so RAFAEL returns only a single answer with the highest score.

An answer consists of the following elements: an answer string, a supporting sentence, a supporting document and a confidence value (the score). A sentence and a document, in which the best mention appeared, are assumed to support the answer. Thanks to properties of Jaccard similarity, the mention score ranges between 0 for completely unrelated  sentences to 1 for practically (ignoring inflection and a word order) the same. Therefore, it may serve as an answer confidence.

When no entity mentions satisfying constraints of a question are found, no answer is returned. This type of result could also be used when the best confidence score is below a predefined value; performance of such technique are shown in section \ref{experiments}. The refusal to answer in case of insufficient confidence plays an important role in \textit{Jeopardy!}, hence in \textit{IBM Watson} \citep{Ferrucci2010}, but it was also used to improve precision in other QA systems \citep{Oh2013}.

\section{Deep Entity Recognition}
\label{DeepER}

Deep Entity Recognition procedure is an alternative to applying Named Entity Recognition in QA to find entities matching question constraints. It scans a text and finds words and multi-word expressions, corresponding to entities. However, it does not assign them to one of several NE categories; instead, WordNet synsets are used. Therefore, named entities are differentiated more precisely (e.g. monarchs and athletes) and entities beyond the classical NE categories (e.g. species, events, devices) could also be recognised in a text.

It does not seem possible to perform this task relying solely on features extracted from words and surrounding text (as in NER), so it is essential to build an entity library. Such libraries already exist ({\em Freebase}, {\em BabelNet}, {\em DBpedia} or {\em YAGO}) and could provide an alternative for DeepER, but they concentrate on English. The task of adaptation of such a base to another language is far from trivial, especially for Slavonic languages with complex NE inflection \citep{Przepiorkowski2007}. An ontology taking into account Polish inflection (\textit{Prolexbase}) has been created by \citet{Savary2013}, but it contains only 40,000 names, grouped into 34 types. 

\subsection{Related work}

The idea of DeepER in a nutshell is to improve QA by annotating a text with WordNet synsets using an entity base created by understanding definitions found in encyclopaedia. Parts of this concept have already appeared in the NLP community.

A technique of coordinating synsets assigned to a question and a possible answer emerged in a study by \citet{Mann2002}. While a question analysis there seems very similar to this work, entity library (called {\em proper noun ontology}) generation differs a lot. The author analysed 1 GB of newswire text and extracted certain expressions, e.g. \textit{"X, such as Y"} implies that Y is an instance of X. Albeit precision of resulting base was not very good (47 per cent for non-people proper names), it led to a substantial improvement of QA performance.

The idea of analysing encyclopaedic definitions to obtain this type of information already appeared, but was employed for different applications. For example, \citet{Toral2006} described a method of building a gazetteer by analysing hyperonymy branches of nouns of first sentences in Wikipedia definitions. Unlike in this work, an original synset was replaced by a coarse-grained NER category. Another example of application is a NE recognizer \citep{Kazama2007} using words from a definition as additional features for a standard CRF classifier. In their definition analysis only the last word of the first nominal group was used.

Other researchers dealt with a task explicitly defined as classifying Wikipedia entries to NER categories. For example \citet{Dakka2008} addressed the problem by combining traditional text classification techniques (bag of words) with contexts of entity mentions. Others \citep{Ponzetto2007} thoroughly examined article categories as a potential source of is-a relations in a taxonomy (99 per cent of entries have at least one category). Inhomogeneity of categories turned out as the main problem, dealt with by a heuristic classifier, assigning is-a and not-is-a labels. Categories were also used as features in a NER task \citep{Richman2008}, but it required a set of manually designed patterns to differentiate between categories of different nature.

Exploring a correspondence between Wikipedia entries and WordNet synsets found an application in automatic enriching ontologies with encyclopaedic descriptions \citep{Ruiz-Casado2005}. However, only NEs already appearing in the WordNet were considered. The task (solved by bag-of-words similarity) is non-trivial only in case of polysemous words, e.g. which of the meanings of {\em Jupiter} corresponds to which Wikipedia article? Others \citep{Toral2008} concentrated on the opposite, i.e. extending the WordNet by NEs that are {\em not} there yet by adding titles of entries as instances of synsets corresponding to their common category.

Also, some see Wikipedia as an excellent source of high-quality NER training data. Again, it requires to project entries to NE categories. A thorough study of this problem, presented by \citet{Balasuriya2009}, utilizes features extracted from article content (bag of words), categories, keywords, inter-article and inter-language links. A final annotated corpus turns out as good for NER training as a manually annotated gold standard.

Finally, some researchers try to generalise NER to other categories, but keep the same machine-learning-based approach. For example, \citet{Ciaramita2006} developed a tagger, assigning words in a text to one of 41 \textit{supersenses}. Supersenses include NE categories, but also other labels, such as \textit{plant}, \textit{animal} or \textit{shape}. The authors projected word-sense annotations of publicly available corpora to supersenses and applied perceptron-trained Hidden Markov Model for sequence classification, obtaining precision and recall around 77 per cent.

\subsection{Entity Library}
\label{Library}

An entity library for DeepER contains knowledge about entities that is necessary for deep entity recognition. Each of them consists of the following elements (entity \#9751, describing the Polish president, Bronisław Komorowski):
\begin{itemize}
\item Main name: \textit{Bronisław Komorowski},
\item Other names (aliases): \textit{Bronisław Maria Komorowski}, \textit{Komorowski},
\item Description URL: \url{http://pl.wikipedia.org/wiki/?curid=121267},
\item plWordNet synsets: 
\begin{itemize}
\item \textit{<podsekretarz1, podsekretarz stanu1, wiceminister1>} (vice-minister, undersecretary),
\item \textit{<wicemarszałek1>} (vice-speaker of the Sejm, the Polish parliament),
\item \textit{<polityk1>} (politician),
\item \textit{<wysłannik1, poseł1, posłaniec2, wysłaniec1, posłannik1>} (member of a parliament),
\item \textit{<marszałek1>} (speaker of the Sejm),
\item \textit{<historyk1>} (historian),
\item \textit{<minister1>} (minister),
\item \textit{<prezydent1, prezydent miasta1>} (president of a city, mayor).
\end{itemize}
\end{itemize}
A process of entity library extraction is performed offline, before question answering. The library built for deep entity recognition in RAFAEL, based on the Polish Wikipedia (857,952 articles, 51,866 disambiguation pages and 304,823 redirections), contains 809,786 entities with 1,169,452 names (972,592 unique). The algorithm does not depend on any particular feature of Wikipedia, so any corpus containing entity definitions could be used.

\begin{figure}
  \centering
    \includegraphics[width=13.5cm]{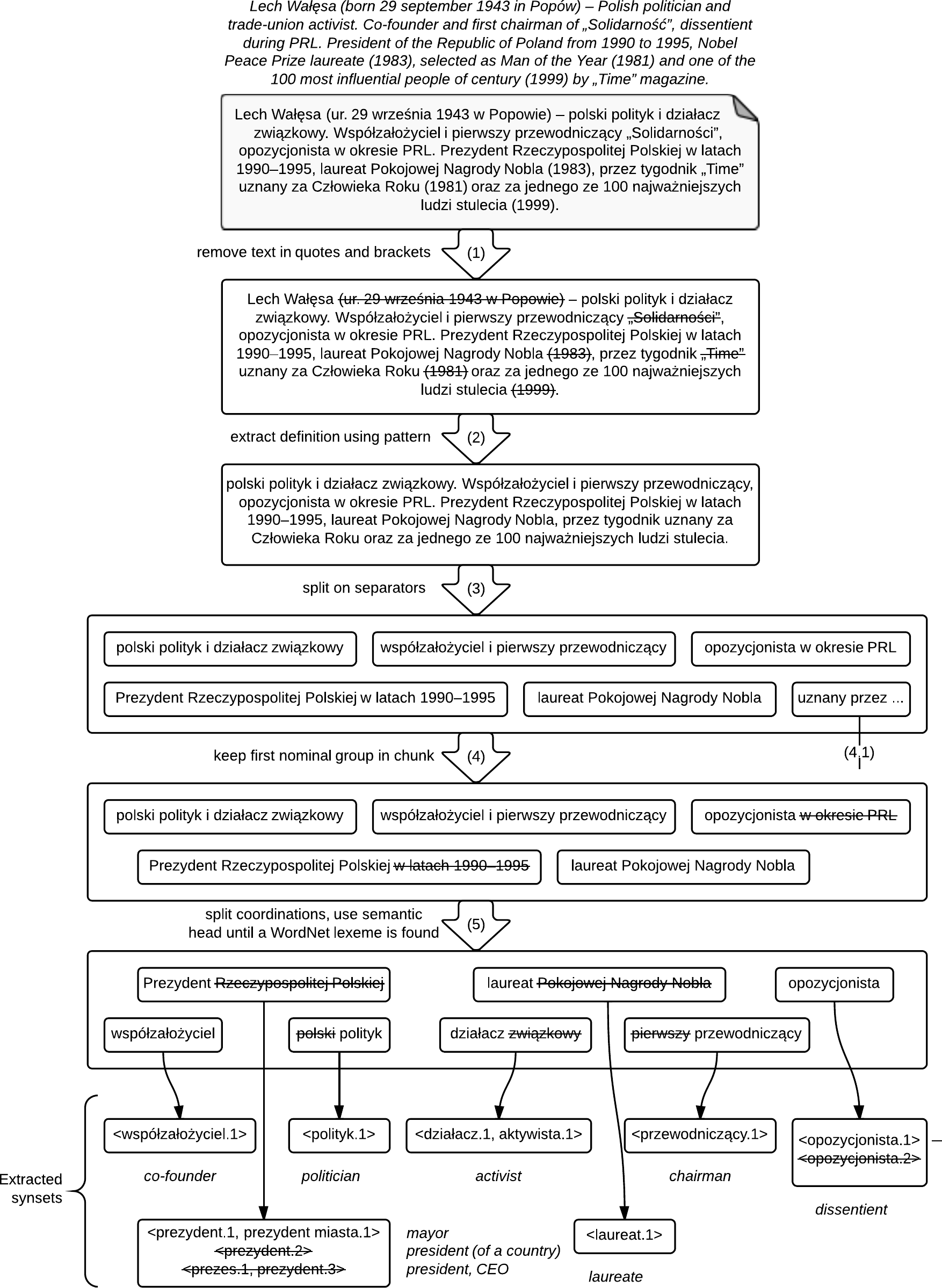}
  \caption{Example of the entity extraction process in DeepER, transforming a Wikipedia entry of \textit{Lech Wałęsa} into a list of synsets.}
  \label{fig:example}  
\end{figure}

Figure \ref{fig:example} shows an exemplary process of converting the first paragraph of a Polish Wikipedia entry, describing former Polish president Lech Wałęsa\footnote{\myurl}, into a list of WordNet synsets. First, we omit all unessential parts of the paragraph (1). This includes text in brackets or quotes, but also introductory expressions like \textit{jeden z} (one of) or \textit{typ} (type of). Then, an entity name is detached from the text by matching one of definition patterns (2). In the example we can see the most common one, a dash (–). Next, all occurrences of separators (full stops, commas and semicolons) are used to divide the text into separate chunks (3). The following step employs shallow parsing annotation -- only nominal groups that appear at the beginning of the chunks are passed on (4). The first chunk that does not fulfil this requirement and all its successors get excluded from further analysis (4.1). Finally, we split the coordination groups and check, whether their lemmas correspond to any lexemes in WordNet (5). If not, the process repeats with the group replaced by its semantic head. In case of polysemous words, only the first word sense (usually the most common) is taken into account.

The whole process is more complicated than the simple example shows. Generally, it consists of the following steps:
\begin{description}
\item[Step 0] Prepare a corpus -- data format and annotation process is the same as for a knowledge base, used in question answering, see section \ref{preprocessing}. It differs in scope of page categories, including not only articles, but also disambiguation and redirection pages.
\item[Step 1] For each of article pages, extract the first paragraph and apply {\em readDefinition} function. If a resulting entity has a non-empty synset list, add it to the library. If some of the redirection pages point to the entity name, add their names as entity aliases.
\item[Step 2] For each of disambiguation pages, extract all items and apply {\em readDefinition} function. If an item refers to an existing entity, extend it with extracted synsets and disambiguation page name. Create a new entity otherwise. Add redirection names as previously.
\item[Step 3] Save the obtained base for future use.
\end{description}
 
\begin{algorithm}
\label{readDefinition}
\caption{Function \textbf{readDefinition}($text$) -- interprets a definition to assign synsets to an entity.}
\KwIn{$text$ - annotated first paragraph of an encyclopaedic entry}
\KwOut{$synsets$ - synsets describing an entity}
\Begin{
	$synsets$ := \{\}\;
	$text$ := removeInBrackets($text$)\;
	$text$ := removeInQuotes($text$)\;
	\ForEach{$pattern$ \textbf{in} $definitionPatterns$}{
		\If{$pattern$ matches $text$}{
			$definition$ := match($pattern$,$text$).group(2)\;
			break\;
		}
	}
	$definition$ := removeDefinitionPrefixes($definition$)\;
	$parts$ := split($definition$,$seperators$)\;
	\ForEach{$part$ \textbf{in} $parts$}{
		$chunk$ := firstGroupOrWord($part$)\;
		\If{isNominal($chunk$)}{
			$synsets$ := $synsets$ $\cup$ extractSynsets($chunk$)\;
		}
		\lElse{
			break\;
		}
	}
	\KwRet{$synsets$}\;
}
\end{algorithm}

The {\em readDefinition} function (shown as algorithm \ref{readDefinition}) analyses a given paragraph of text and extracts a set of synsets, describing an entity, to which it corresponds, as exemplified by figure \ref{fig:example}. Simplifying, it is done by removing all unnecessary text (in brackets or quotes), splitting it on predefined separators (commas, full stops, semicolons) and applying \textit{extractSynsets} function with an appropriate stop criterion. The {\em readDefinition} makes use of the following elements:
\begin{description}
\item[\em removeInBrackets()] removes everything that is between brackets ([], () or \{\}) from the text (step (1) in figure \ref{fig:example}). 
\item[\em removeInQuotes()] removes everything between single or double quotes from the text (step (1) in the example).
\item[\em definitionPatterns] contains patterns of strings separating a defined concept from a definition, e.g. hyphens or dashes (used in step (2) of the example) or \textit{jest to} (is a).
\item[\em removeDefinitionPrefixes()] removes expressions commonly prefixing a nominal group, such as \textit{jeden z} (one of), \textit{typ} (a type of) or \textit{klasa} (a class of), not present in the example.
\item[\em separators] a set of three characters that separate parts of a definition: ".", "," and ";".
\item[\em firstGroupOrWord()] returns the longest syntactic element (syntactic group or word) starting at the beginning of a chunk (step (4) in the example).
\item[\em isNominal()] decides, whether a chunk is a noun in nominative, a nominal group or a coordination of nominal groups.
\end{description}

\begin{algorithm}
\label{extractSynsets}
\caption{Function \textbf{extractSynsets}($chunk$) -- recursively extracts synsets from a nominal chunk.}
\KwIn{$chunk$ - a nominal chunk (a syntactic group or a single noun)}
\KwOut{$synsets$ - WordNet synsets corresponding to $chunk$}
\Begin{
	$lemma$ := lemmatise($chunk$)\;
	\If{inWordNet($lemma$)}{
		\KwRet{getLexemes($lemma$).synset(0)}\;
	}
	\ElseIf{isCoordination($chunk$)}{
		$synsets$ := \{\}\;
		\ForEach{$element$ \textbf{in} $chunk$}{
			$synsets$ := $synsets$ $\cup$ extractSynsets($element$)\;
		}
		\KwRet{$synsets$}\;
	}
	\ElseIf{isGroup($chunk$)}{
		\KwRet{extractSynsets($chunk$.semanticHead)}\;
	}
	\lElse{
		\KwRet{\{\}}\;
	}
}
\end{algorithm}

The {\em extractSynsets} function (shown as algorithm \ref{extractSynsets}) accepts a nominal chunk and extracts WordNet synsets, corresponding to it. It operates recursively to dispose any unnecessary chunk elements and find the longest subgroup, having a counterpart in WordNet. It corresponds to step (5) in figure \ref{fig:example} and uses the following elements:
\begin{description}
\item[\em lemmatise()] returns a lemma of a nominal group. 
\item[\em inWordNet()] checks whether a given text corresponds to a lexeme in WordNet.
\item[\em getLexemes()] return a list of WordNet lexemes corresponding to a given text. 
\item[\em synset()] return a synset including a lexeme in a given word sense number.
\item[\em isCoordination()] return TRUE iff a given chunk is a coordination group.
\item[\em isGroup()] return TRUE iff a given chunk is a group.
\item[\em semanticHead] is an element of a syntactic group, denoted as a semantic head.
\end{description}

A few of design decisions reflected in these procedures require further comment. First of all, they differ a lot from the studies that involve a definition represented with a bag of words \citep{Dakka2008,Ruiz-Casado2005,Balasuriya2009}. Here, a certain definition structure is assumed, i.e. a series of nominal groups divided by separators. What is more, as the full stop belongs to them, the series may continue beyond a single sentence, which has improved recall in preliminary experiments. Availability of a shallow parsing layer and group lemmatisation allows to query WordNet by syntactic groups instead of single nouns, as in work of \citet{Toral2006}. As word order is relatively free in Polish, a nominal group cannot be assumed to end with a noun, like \citet{Kazama2007} did. Instead, a semantic head of a group is used.

Finally, the problem of lack of word sense disambiguation remains -- the line \textit{getLexemes($lemma$).synset(0)} means that always a synset connected to the first meaning of a lexeme is selected. We assume that it corresponds to the most common meaning, but that is not always the case -- in our example at figure \ref{fig:example} <prezydent.1, prezydent miasta.1> (president of a city, i.e. mayor) precedes <prezydent.2> (president of a country, the obvious meaning). However, it does not have to harm QA performance as far as the question analysis module (section \ref{analysis}) functions analogously, e.g. in case of a question beginning with \textit{który prezydent\ldots} (which president\ldots). Therefore, the decision has been motivated by relatively good performance of this solution in previously performed experiments on question analysis \citep{Przybya2013}. It also works in other applications, e.g. gazetteers generation \citep{Toral2006}.

To assess quality of the entity library, its content has been compared with synsets manually extracted from randomly selected 100 Wikipedia articles. 95 of them contain a description of an entity in the first paragraph. Among those, DeepER entity library includes 88 (per-entity recall 92.63 per cent). 135 synsets have been manually assigned to those entities, while the corresponding set in library contains 133 items. 106 of them are equal (per-synset precision 79,70 per cent), while 13 differ only by word sense. 16 of manually extracted synsets hove no counterpart in the entity library (per-synset recall 88.15 per cent), which instead includes 14 false synsets.

\subsection{Entity Recognition}
\label{Recognition}

An entity recognition step is performed within the question answering process and aims at selecting all entity mentions in a given annotated document. Before it begins, the entity library is read into a PATRICIA trie, a very efficient prefix tree. In this structure, every entity name becomes a key for storing a corresponding list of entities.

When a document is ready for analysis, it is searched for strings that match any of the keys in the trie. The candidate chunks (sequences of segments) come from three sources:
\begin{enumerate}
\item lemmata of words and syntactic groups,
\item sequences of words in surface forms (as they appear in text),
\item sequences of words in base forms (lemmata).
\end{enumerate}
The last two techniques are necessary, because a nominal group lemmatisation often fails, especially in case of proper names. Their rich inflection in Polish \citep{Przepiorkowski2007} means that a nominal suffix of an entity may be hard to predict. Therefore, a chunk is considered to match an entity name if:
\begin{itemize}
\item they share a common prefix,
\item an unmatched suffix in neither of them is longer than 3 characters,
\item the common prefix is longer than the unmatched chunk suffix.
\end{itemize}
Given a list of entity mentions, RAFAEL checks their compatibility with a question model. Two of its constituents are taken into account: a general question type and a synset. An entity mention agrees with NAMED\_ENTITY type if its first segment starts with a capital letter and always agrees with UNNAMED\_ENTITY. To pass a semantic agreement test, the synset of the question model needs to be a (direct or indirect) hypernym of one of the synsets assigned to the entity. For example, list of synsets assigned to entity \textit{Jan III Sobieski} contains <król.1> (king), so it matches a question focus <władca.1, panujący.1, hierarcha.2, pan.1> (ruler) through a hypernymy path <władca.1, panujący.1, hierarcha.2, pan.1> $\rightarrow$ <monarcha.1, koronowana głowa.1> (monarch) $\rightarrow$ <król.1>. All the mentions of entities satisfying these conditions are returned for further processing. 

\section{Evaluation}
\label{Evaluation}

Evaluation of RAFAEL is typical for factoid QA systems: given a knowledge base and and questions, its  responses are compared to the expected ones, prepared in advance. Section \ref{data} describes data used in this procedure, whereas section \ref{automatic} explains how an automatic evaluation is possible without human labour.

\subsection{Data}
\label{data}

The Polish Wikipedia serves as a knowledge base. It has been downloaded from a project site as a single database dump at 03.03.2013, from which plain text files have been extracted using \textit{Wikipedia Extractor} 2.2 script\footnote{\url{http://medialab.di.unipi.it/wiki/Wikipedia_Extractor}}. It means that only plain text is taken into account -- without lists, infoboxes, tables, etc. This procedure leads to a corpus with 895,486 documents, containing 168,982,550 segments, which undergo the annotation process, described in section \ref{preprocessing}.

The questions that are to be answered with the knowledge base come from two separate sets:
\begin{enumerate}
\item {\em Development set} bases on 1500 (1130 after filtering\footnote{The questions that were filtered out either do not belong to \textit{factoid} category, as defined at the beginning of chapter \ref{RAFAEL} (e.g. demand calculations, translations or long explanations) or lack answer in  the Polish Wikipedia. }) questions from a Polish quiz TV show, called \textit{Jeden z dziesięciu} \citep{Karzewski1997}. It was involved in previous experiments \citep{Przybyla,Przybya2013}. 
\item {\em Evaluation set} bases on an open dataset for Polish QA systems, published by \citet{Marcinczuk2013}. It has been gathered from \textit{Did you know\ldots} column, appearing in the main page of the Polish Wikipedia. It contains 4721 questions, from which 1000 have been analysed, which resulted in 576 satisfying the task constrains, given in chapter \ref{RAFAEL}. 
\end{enumerate}
Table \ref{tab:types} shows a distribution of different question types and named entity types in the sets.

\begin{table}
\begin{tabular}{  r  l  c  c  }
\hline
\hline
\textbf{General type} & \textbf{Named entity type} & \textbf{Development} & \textbf{Final evaluation}\\
\hline
WHICH & - & 2.39\% & 0.17\% \\
TRUEORFALSE & - & 2.21\% & 0.87\% \\
MULTIPLE & - & 2.57\% & 8.33\% \\
OTHER NAME & - & 2.04\% & 0.87\% \\
UNNAMED\_ENTITY & - & 32.30\% & 16.67\% \\
\hline
\multirow{31}{*}{NAMED\_ENTITY} & PLACE & 2.83\% & 9.03\% \\
& CONTINENT & 0.35\% & 0.17\% \\
& RIVER & 0.97\% & 0.17\% \\
& LAKE & 0.80\% & 0.00\% \\
& MOUNTAIN & 0.35\% & 0.52\% \\
& RANGE & 0.18\% & 0.17\% \\
& ISLAND & 0.44\% & 0.17\% \\
& ARCHIPELAGO & 0.18\% & 0.00\% \\
& SEA & 0.18\% & 0.00\% \\
& CELESTIAL\_BODY & 0.71\% & 0.17\% \\
& COUNTRY & 4.60\% & 0.35\% \\
& STATE & 0.62\% & 0.17\% \\
& CITY & 4.69\% & 0.35\% \\
& NATIONALITY & 1.06\% & 0.17\% \\
& PERSON & 22.92\% & 36.81\% \\
& NAME & 0.97\% & 0.00\% \\
& SURNAME & 0.88\% & 0.00\% \\
& BAND & 0.53\% & 0.17\% \\
& DYNASTY & 0.53\% & 0.17\% \\
& ORGANISATION & 1.86\% & 3.47\% \\
& COMPANY & 0.18\% & 0.52\% \\
& EVENT & 0.97\% & 1.39\% \\
& TIME & 0.18\% & 3.82\% \\
& CENTURY & 0.80\% & 0.00\% \\
& YEAR & 3.01\% & 0.69\% \\
& PERIOD & 0.09\% & 0.17\% \\
& COUNT & 2.74\% & 7.64\% \\
& QUANTITY & 0.53\% & 1.74\% \\
& VEHICLE & 0.88\% & 1.74\% \\
& ANIMAL & 0.09\% & 0.00\% \\
& TITLE & 3.36\% & 3.47\% \\
\hline
\hline
\end{tabular}
  \caption{A distribution of different general types and named entity types in development (1130 questions) and final evaluation (576 questions) sets.}
  \label{tab:types}
\end{table}

To each of the questions from both sets some information has been assigned manually. It includes an identification number, an expected answer string, a general question type, a named entity type (if applicable) and an expected source document. Table \ref{tab:questions} contains several exemplary questions from the development set.

\begin{table}
\begin{tabular}{  c  c  c }
\hline
\hline
\multicolumn{3}{c}{\textbf{Question}}\\
\textbf{Question type}&\textbf{Source article}&\textbf{Answer}\\
\hline
\multicolumn{3}{c}{ W którym roku umarł Stefan Żeromski?}\\
\multicolumn{3}{c}{ \textit{What year did Stefan Żeromski die?}}\\
NAMED\_ENTITY & \multirow{2}{*}{Stefan Żeromski} & \multirow{2}{*}{1925} \\
:YEAR &  &  \\
\hline
 \multicolumn{3}{c}{Jakie organella nadają barwę korzeniom marchwi?}\\
\multicolumn{3}{c}{\textit{What organelles does the carrot take its colour from?}}\\
  \multirow{2}{*}{UNNAMED\_ENTITY} & \multirow{2}{*}{Chromoplast} & Chromoplasty \\
  &  & \textit{Chromoplasts} \\
  \hline
  \multicolumn{3}{c}{Jakiego wyznania jest większość mieszkańców Liechtensteinu?}\\
\multicolumn{3}{c}{\textit{What is the major confession in Liechtenstein?}}\\
  \multirow{2}{*}{UNNAMED\_ENTITY} & \multirow{2}{*}{Liechtenstein} & Katolicyzm \\
    &  & \textit{Catholicism} \\
    \hline
  \multicolumn{3}{c}{Który z filozofów był twórcą „atomizmu”?}\\
\multicolumn{3}{c}{\textit{Which philosopher formulated the atomic theory?}}\\
  NAMED\_ENTITY & \multirow{2}{*}{Demokryt} & Demokryt z Abdery  \\
   :PERSON &  & \textit{Democritus of Abdera} \\
   \hline
  \multicolumn{3}{c}{Czy Jacques Brel pochodził z Francji?}\\
\multicolumn{3}{c}{\textit{Was Jacques Brel born in France?}}\\
 \multirow{2}{*}{TRUEORFALSE} & \multirow{2}{*}{Jacques Brel} & Nie  \\
   &  & \textit{No} \\
  \hline
  \hline
\end{tabular}
  \caption{Exemplary questions with their types (general and named entity), expected source articles and answers.}
  \label{tab:questions}
\end{table}

The additional information (question types and expected documents) makes it possible to evaluate only selected modules of the whole QA system. For example, we could test question classification by comparing results against given question types or entity selection by analysing only the relevant document.

\subsection{Automatic Evaluation}
\label{automatic}

Thanks to availability of the DeepER entity library, it is possible to automatically perform answer evaluation for all the question types that are recognised by this technique (UNNAMED\_ENTITY and NAMED\_ENTITY excluding dates, numbers and quantities).

Both an expected and obtained answer are represented as short strings, e.g. \textit{Bronisław Komorowski}. However, it does not suffice to check their exact equality. That is caused by existence of different names for one entity (\textit{Bronisław Maria Komorowski} or \textit{Komorowski}), but also rich nominal inflection (\textit{Komorowskiego}, \textit{Komorowskiemu}, \ldots).

In fact, we want to compare {\em entities}, not {\em names}. Hence, deep entity recognition is a natural solution here. To check correctness of an answer, we use it as an input for the recognition process, described in section \ref{Recognition}. Then, it is enough to check whether the expected answer appears in any of lists of names, assigned to the recognized entities. For example, let us consider a question: \textit{Kto jest obecnie prezydentem Polski?} (Who is the current president of Poland?) with expected answer \textit{Bronisław Komorowski} and a system answer \textit{Komorowski}. The DeepER process finds many entities in the string (all the persons bearing this popular surname). One of them is the question goal, hence, has \textit{Bronisław Komorowski} in its list of names.

As the process of entity recognition is imperfect, so is the automatic evaluation. However, it still lets us to notice general trends in answering performance with respect to several factors. Of course, the final evaluation needs to be checked manually. 

\section{Results}

As mentioned in previous section, the results consist of two groups: experiments, showing an influence of some aspects of algorithm on performance, and a final assessment. Both use the Polish Wikipedia as a knowledge base, whereas the questions asked belong to development and evaluation sets, respectively. In this section, {\em recall} measures percentage of questions, to which RAFAEL gave any answer, whereas {\em precision} denotes percentage of question answered correctly.

When analysing results of different entity recognition techniques, we need to remember that they strongly rely on output of the question analysis, which is not perfect. In particular, tests show that 15.65 per cent of questions is assigned to wrong type and 17.81 per cent search results do not include the expected document \citep{Przybya2013}. The entity recognition (ER) stage, a focus of this work, is very unlikely to deliver valid answers in these cases. However, as the expected question type and source document are available in question metadata, it is possible to {\em correct} results of question analysis by artificially replacing a wrong type and/or adding the expected document to the retrieved set. In that way the ER modules could be evaluated, as if question analysis worked perfectly. Note that this approach slightly favours NER-based solutions as the question metadata contains general types and named entity types but lack focus synsets, used by DeepER.

\subsection{Experiments}
\label{experiments}

\begin{figure}
  \centering
    \includegraphics[width=13.5cm]{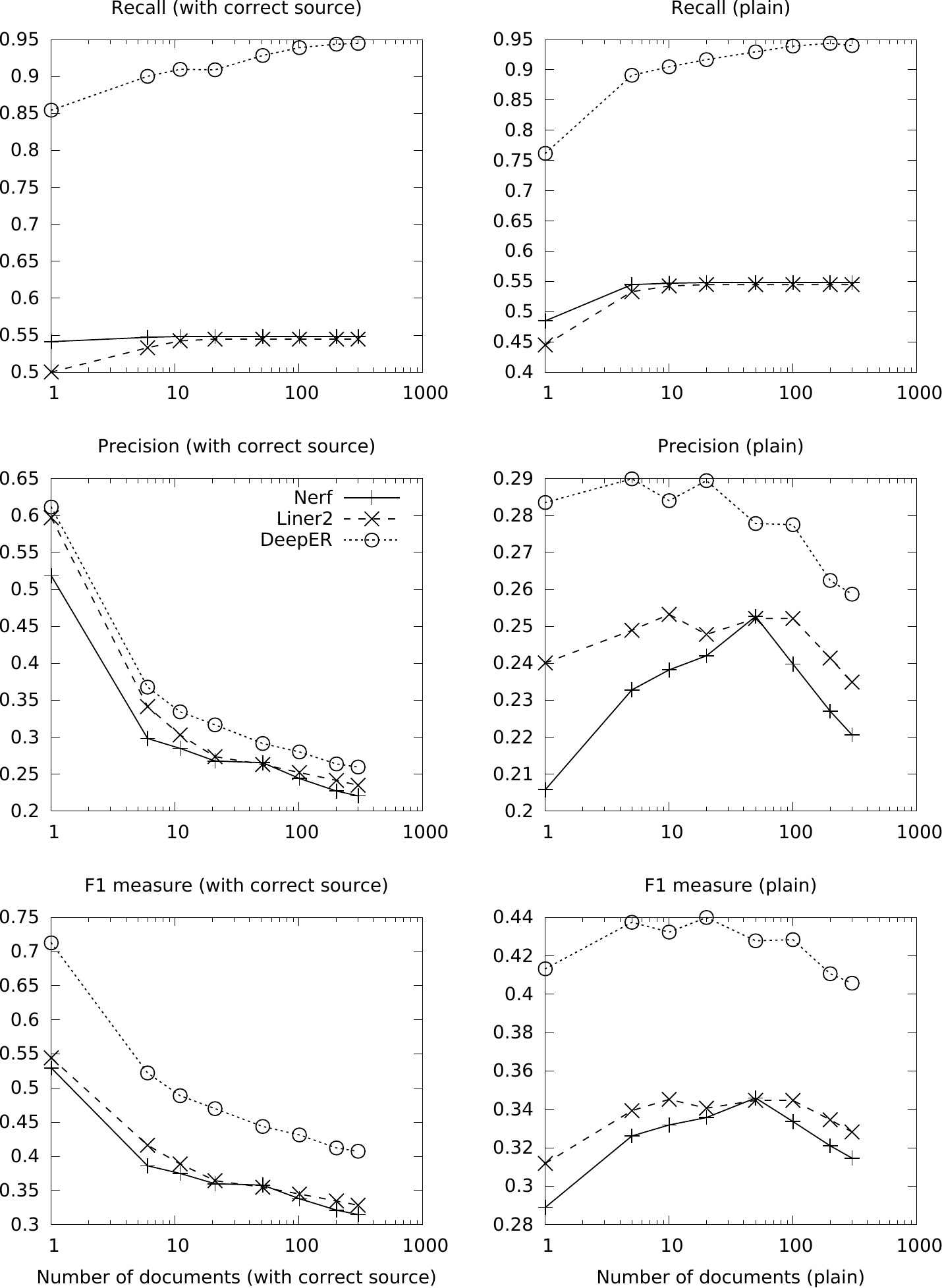}
  \caption{Question answering performance with respect to size of a retrieved set of documents, undergoing a full analysis. Two versions are considered -- with and without guaranteed presence of an article, containing the desired information, in a set. The results for different entity recognition techniques-- traditional NER ({\em Nerf}, {\em Liner2}) and {\em DeepER}.}
  \label{fig:plot1}  
\end{figure}
The goal of the first experiment is to test how number a of documents retrieved from the search engine and analysed by the entity recognition techniques, influences the performance. Question classification errors have been bypassed as described in the previous paragraph. Additionally, two versions have been evaluated: with and without corrections of a retrieved set of documents. Figure \ref{fig:plot1} demonstrates results for different entity recognition techniques.  

As we can see, if a retrieved set contains the desired article, adding new documents slightly increases recall, while precision drops observably. That is because additional irrelevant documents usually introduce noise. However, in some cases they are useful, as increasing recall indicates. On the other hand, if we have no guarantee of presence of the expected document in a list, it seems more desirable to extend it, especially for small sizes. For sets bigger than 50 elements, the noise factor again dominates our results. Judging by F1 measure, the optimal value is 20 documents.

When it comes to the comparison, it should be noted that DeepER performs noticeably better than traditional NER. The gain in precision is small, but recall is almost twice as big. It could be easily explained by the fact that the NER solutions are unable to handle UNNAMED\_ENTITY type, which accounts for 36 per cent of the entity questions.

\begin{figure}
  \centering
    \includegraphics[width=7cm]{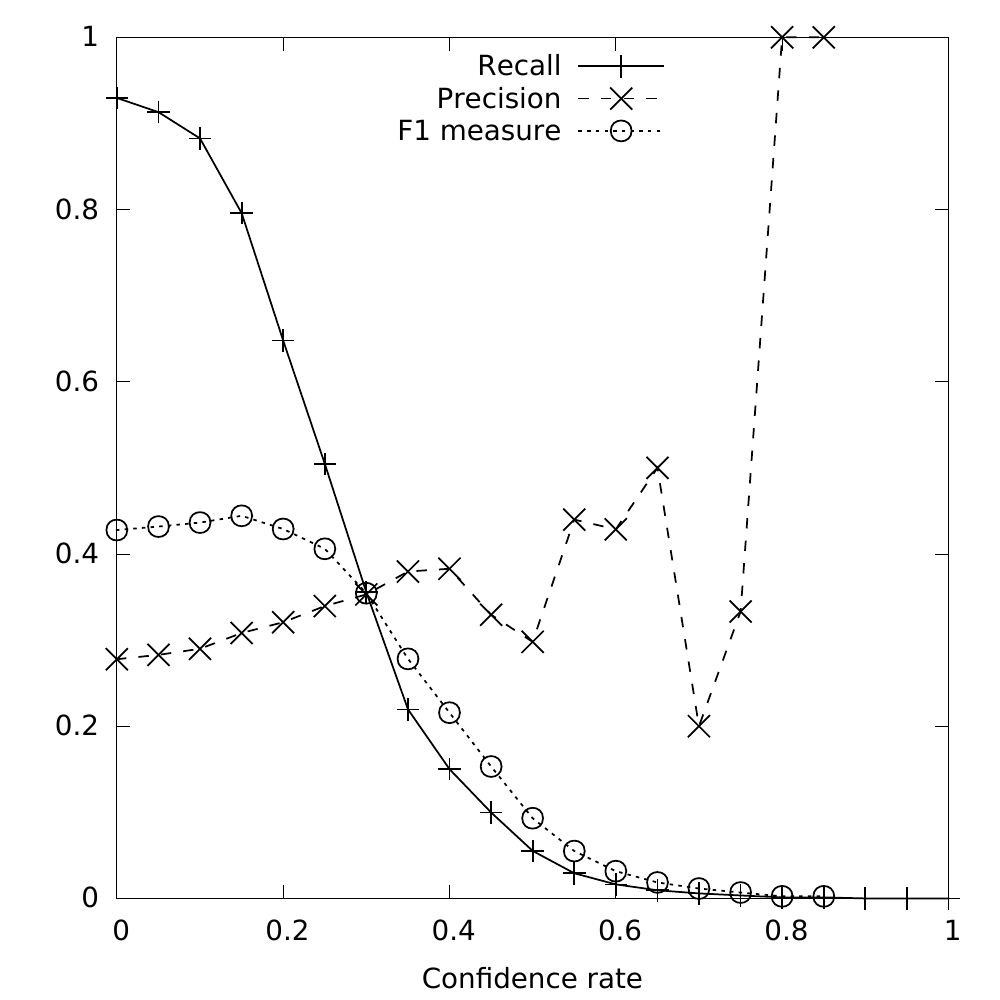}
  \caption{RAFAEL performance with respect to minimal confidence rate. Results computed using DeepER with corrected question type and corrected list of 50 documents.}
  \label{fig:plot2}  
\end{figure}
It is also worthwhile to check how the system performs while using different values of minimal confidence rate (Jaccard similarity), as described in section \ref{generation}. It could become useful when we demand higher precision and approve lower recall ratio. The plot in figure \ref{fig:plot2} shows answering performance using DeepER with corrected question analysis with respect to the minimal confidence rate. Generally, the system behaves as expected, but the exact values disappoint. The precision remain at a level of 25-40 per cent up to confidence 0.75, where in turn recall drops to 0.35 per cent only. Values of F1 measure suggest that 0.2 is the highest sensible confidence rate.

\begin{figure}
  \centering
    \includegraphics[width=10cm]{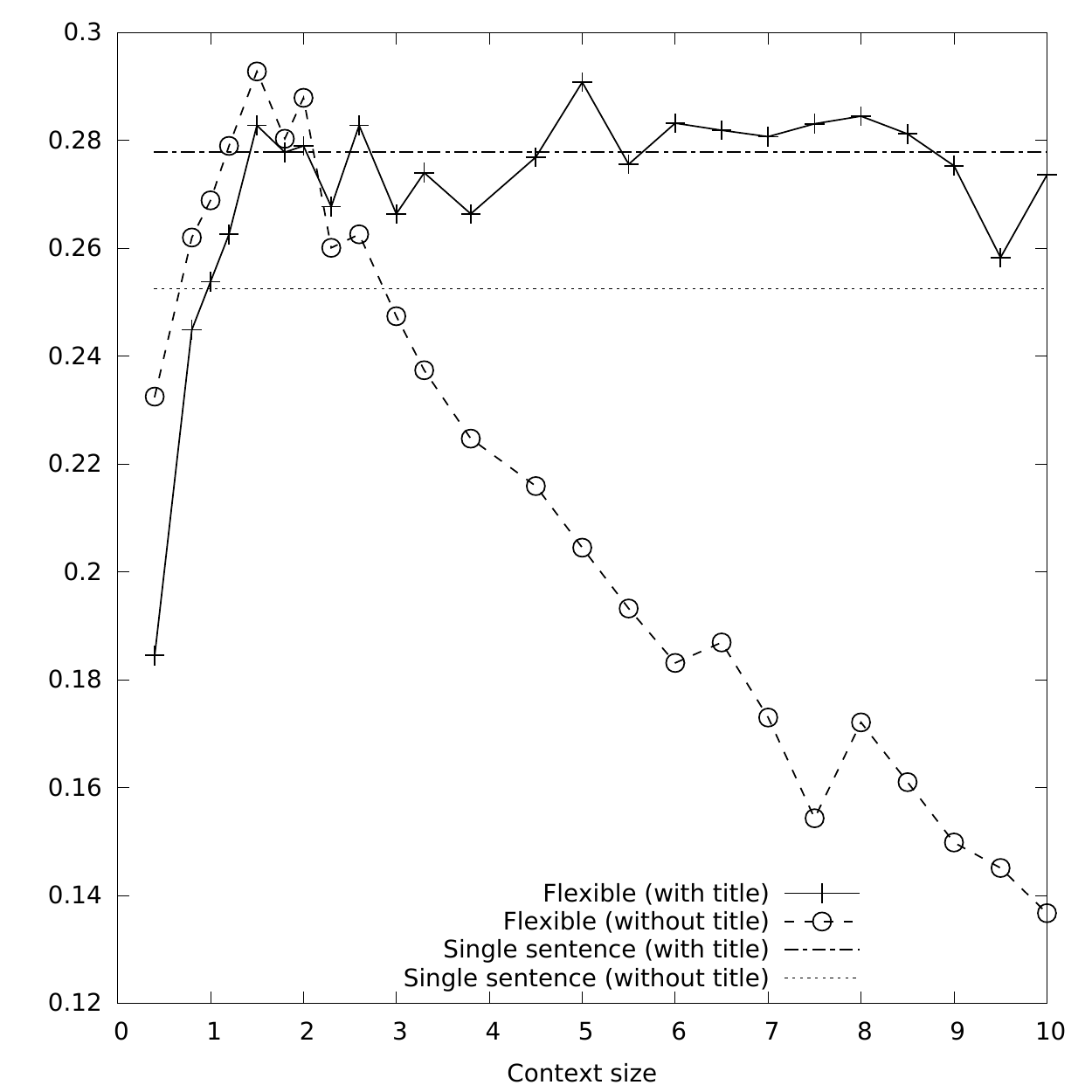}
  \caption{Question answering performance for different context generation strategies: single sentence and sequence of segments of certain length. Both types considered with and without an article title added.}
  \label{fig:plot3}  
\end{figure}
One more parameter worth testing, explained in section \ref{context}, is the context generation strategy. To find the entity with a context most similar to a question content, we could analyse a single sentence, where it appears, or a sequence of words of a predefined length\footnote{Expressed as a multiplication of question content size.}. For both of these solutions, we could also add a document title, as it is likely to be referred to by anaphoric expressions. Figure \ref{fig:plot3} shows the value of precision (recall does not depend on context) for these four solutions.

We can see that inclusion of a title in a context helps to achieve a better precision. The impact of anaphoric reference to title emerges clearly in case of flexible context -- the difference grows with context size. Quite surprisingly, for the optimal context length (1.5 * question size), it is on the contrary. However, because of the small difference between the techniques including title, for the sake of simplicity, the single sentence is used in the final evaluation.

\subsection{Final System Evaluation}
\label{final}
To impose a realistic challenge to the system, the evaluation set, used at this stage, substantially differs from the one used during the development (see section \ref{data}). A configuration for the final evaluation has been prepared based on results of the experiments. All of the tested versions share the following features:
\begin{itemize}
\item no question analysis corrections,
\item question classification and query generation solutions which proved best in the previous experiments (see section \ref{analysis}),
\item a retrieved set of documents including 20 articles,
\item no minimal confidence,
\item singe sentence context with title.
\end{itemize}

Tested solutions differ with respect to entity recognition only; RAFEL variants based on the following options are considered: 
\begin{itemize}
\item quantities recognizer (\textit{Quant}),
\item traditional NER solutions: \textit{Nerf} and \textit{Liner2},
\item deep entity recognition (\textit{DeepER}),
\item hybrid approach, where entity mentions were gathered from all the above sources.
\end{itemize}

\begin{table}
\begin{tabular}{ r  c  c c  c  c  c  c  c  c }
\hline
\hline
&\multicolumn{2}{c}{\textbf{Recall}}&\multicolumn{3}{c}{\textbf{Precision}}&\multicolumn{2}{c}{\textbf{F1 measure}}&\multicolumn{2}{c}{\textbf{MRR}}\\
\textbf{ER}&&&\multicolumn{2}{c}{\textbf{manual}} &\textbf{auto} &&&&\\
\textbf{solution}&\textbf{value}&$\boldsymbol{\sigma}$&\textbf{value}&$\boldsymbol{\sigma}$&\textbf{value}&\textbf{value}&$\boldsymbol{\sigma}$&\textbf{value}&$\boldsymbol{\sigma}$\\
\hline
Quant & 9.55\% & 1.19\% & 27.27\% & 5.6\% & -& 0.1415 & 0.0162 & 26.71\% & 4.86\% \\
Nerf & 56.25\% & 2.12\% & 34.88\% & 2.73\% & -& 0.4306 & 0.0213 & 33.66\% & 2.29\%\\
Liner2 & 45.31\% & 2.05\% & 39.08\% & 2.90\% &  34.10\% & 0.4197 & 0.0188 & 41.36\% & 2.70\%\\
DeepER & 72.92\% & 1.88\% & 35.24\% & 2.23\% & 33.89\%& 0.4751 & 0.0214 & 32.80\% & 1.99\%\\
Hybrid & 89.58\% & 1.24\% & 33.14\% & 2.01\% & -& 0.4838 & 0.0221 & 35.57\% & 1.88\%\\
\hline
\hline
\end{tabular}
  \caption{Question answering accuracy of RAFAEL with different entity recognition strategies: quantities only (\textit{Quant}), traditional NER (\textit{Nerf}, \textit{Liner2}), deep entity recognition (\textit{DeepER}) and their combination (\textit{Hybrid}).}
  \label{tab:final}
\end{table}
Table \ref{tab:final} shows results of the final evaluation, expressed by recall, precision, F1 measure and Mean Reciprocal Rank (MRR)\footnote{Evaluated automatically due to length of ranking lists.}. Standard deviations of these values have been obtained by bootstrap resampling of the test set. Additionally, precision obtained by automatic evaluation has been added, where applicable\footnote{Results obtained using \textit{Quant} and \textit{Nerf} may be expressed by quantities and dates, which are not supported by the entity-based automatic evaluation}. As we can see, only a small percentage of questions is handled by the quantitative entities recognition. NER-based solutions deal with slightly more (\textit{Nerf}) or less (\textit{Liner2}) than a half of the questions. When using DeepER, the recall ratio rises to 73 per cent while the precision does not differ significantly. That is because UNNAMED\_ENTITY questions (unreachable for traditional NER) account for a substantial part of the test set. The maximum recall is obtained by the hybrid solution (90 per cent) but it comes at a cost of lower precision (33 per cent). On the other hand,  when we take the whole ranking lists into account, traditional NERs seem to perform better (in terms of MRR).

As expected, the automatic evaluation underestimates precision, but the difference remains below 5 per cent. Judging by F1 measure, the hybrid solution seems to beat the others.

\section{Discussion}
\label{discussion}

The main strength of DeepER compared to NER, according to results shown in figure \ref{tab:final}, is much higher recall. Table \ref{tab:recall} shows examples of questions, to which only DeepER provides a correct answer. As we can see (notice question foci in the table), they could not be assigned to any of the traditional NE categories.

\begin{table}
\begin{tabular}{c c}
\hline
\hline
\multicolumn{2}{c}{\textbf{Question}}\\
\textbf{Question focus}&\textbf{Answer}\\
\hline
\multicolumn{2}{c}{Który żleb uważany jest za najbardziej lawiniasty w całych Karpatach?}\\
\multicolumn{2}{c}{\textit{Which gulley is considered the most avalanche-prone in the Carpathians?} }\\
  żleb & \multirow{2}{*}{Pusty Żleb} \\
  \textit{gulley}  &  \\
\hline
\multicolumn{2}{c}{Jaki owad pożera liście owadożernej rosiczki?}\\
\multicolumn{2}{c}{\textit{What insect feeds on leaves of carnivorous sundew?}}\\
  owad & Piórolotek bagniczek \\
   \textit{insect} & \textit{Buckleria paludum} \\
\hline
\multicolumn{2}{c}{Jaki przyrząd pozwala na pomiar współczynnika załamania światła?}\\
\multicolumn{2}{c}{ \textit{What apparatus is used to measure the refractive index?}}\\
  przyrząd  & Refraktometr Abbego \\
 \textit{apparatus} &  \textit{Abbe refractometer} \\
\hline
\multicolumn{2}{c}{Jaki związek chemiczny służy do otrzymywania boru o wysokiej czystości?}\\
\multicolumn{2}{c}{ \textit{What chemical compound is used to obtain boron of high purity?}}\\
  związek chemiczny & Jodek boru  \\
  \textit{chemical compound} & \textit{Boron triiodide} \\
\hline
\multicolumn{2}{c}{Jaką metodą łamano szyfry Enigmy przed wynalezieniem cyklometru?}\\
\multicolumn{2}{c}{ \textit{What method was used to decrypt Enigma cipher before the advent of the cyclometer?}}\\
    metoda  & Metoda rusztu  \\
  \textit{method}  & \textit{Grill} \\
\hline
\hline
\end{tabular}
  \caption{Examples of questions which have been handled and answered correctly only with the DeepER approach. Their foci lie beyond areas covered by the NE categories.}
  \label{tab:recall}
\end{table}

The other striking fact in the results is low precision. A part of the wrong answers was inspected and most of the errors seem to result from the following phenomena:
\begin{itemize}
\item Information given in question in concise form, in article appears scattered over many sentences.
\item Some of the pairs of words in question content and answer context are very related morphologically or semantically, but differ in lemmata, e.g. \textit{łowca} (huntsman) and \textit{łowiectwo} (huntsmanship) or \textit{myśliwy} (huntsman).
\item Because of rich inflection in Polish, a tagger quite often fails to select the correct lemma of a word, especially in case of proper names.
\item If several entities of a desired type appear in one sentence, the bag-of-words model does not suffice to select the best one.
\end{itemize}
The entity recognizers also introduce errors typical for them:
\begin{itemize}
\item \textit{Quant} ignores quantity types; therefore, a year is frequently returned as an answer to questions beginning with \textit{How many \ldots}.
\item \textit{Nerf} and \textit{Liner2} have insufficient recall -- they recognize only a fraction of entities of a desired type.
\item \textit{DeepER} suffers from lack of word sense disambiguation, which impedes WordNet-based inference.
\end{itemize}
The last remark applies also to other techniques. For example, consider a word \textit{kot}, which means \textit{a cat}. However, it is also a name of a journal, a lake, a village, a badge (\textit{KOT}), a surname of 10 persons in the Polish Wikipedia and much more. A human would usually assume the most common meaning (a cat), but the system treats them as equally probable. It introduces noise in the process, as such an entity matches many types of questions.

Another thing that demands explanation is a difference in precision of answers found using \textit{Liner2} and DeepER: in evaluation set the latter does not maintain its advantage from development set. It could be explained by different compositions of the question sets (table \ref{tab:types}) -- the development one contains much more questions beginning with ambiguous pronouns, followed by a question focus, e.g. \textit{Który poeta\ldots} (which poet), thus providing a precise synset (a poet) for deep entity recognition. Members of the evaluation set much more frequently begin with pronouns like \textit{Kto \ldots}(who), where a synset corresponds to a general NE type (a person).

As RAFAEL is the first Polish QA system, able to answer by entities instead of documents, we can not compare it directly to any other solution. However, the evaluation set has been created based on questions published by \citet{Marcinczuk2013} and used for evaluation of a document retrieval system \citep{Radziszewski2013}. Their baseline configuration achieved a@1 (percentage of questions answered by the first document, corresponds to precision in table \ref{tab:final}) equal 26.09 per cent. By taking into account proximity of keyword matches (MCSW method), they improved the result to 38.63 per cent. We can see that RAFAEL, despite solving much more challenging problem, in all configurations obtains better precision than baseline; using \textit{Liner2} it beats even the best method tested on this set (MCSW).

The results suggest two possible directions of future work to improve performance of RAFAEL. Firstly, involving semantics in sentence matching could solve some of the problems mentioned above. There are a lot of techniques in that area, also in QA systems (see a variety of them used by \citet{Yih2013}), but their implementation in a morphologically rich language would require a thorough study. For example, there exist techniques computing a semantic similarity based on a WordNet graph \citep{Moldovan2002a}, which is available for Polish and proved very useful in this study. Secondly, the relatively good performance of hybrid ER indicates that it may be good to apply different entity recognizer to different questions. For example, we could evaluate them for each question type separately and select the one that performs best for a given one. However, it would require much more training data to have a substantial number of questions of each type, including the scarce ones (observe sparsity of table \ref{tab:types}).

When it comes to DeepER, word ambiguity seem to be the main issue for future efforts. Of course, a full-lexicon precise word-sense disambiguation tool would solve the problem, but we can't expect it in near future. Instead, we could select a synset somewhere in a path between a focus synset and a named entity type. In the example from figure \ref{fig:example} rather than choosing between <prezydent.1, prezydent miasta.1> (president of a city) and <prezydent.2> (president of a country) we could use <urzędnik.1, biuralista.1> (official), which covers both meanings.

\section{Conclusions}
\label{conclusions}

This paper introduces RAFAEL, a complete open-domain question answering system for Polish. It is capable of analysing a given question, scanning a large corpus and extracting an answer, represented as a short string of text.

In its design, the focus has been on entity recognition techniques, used to extract all the entities compatible with a question from a given text. Apart from the traditional named entity recognition, differentiating between several broad categories of NEs, a novel technique, called Deep Entity Recognition (DeepER), has been proposed and implemented. It is able to find entities belonging to a given WordNet synset, using an entity library, gathered by interpreting definitions from encyclopaedia.

Automatic evaluation, provided by DeepER approach, has let to perform several experiments, showing answering accuracy with respect to different parameters. Their conclusions have been used to prepare final evaluation, which results have been checked manually. They suggest that the DeepER-based solution yields similar precision to NER, but is able to answer much more questions, including those beyond the traditional categories of named entities.

\section*{Appendix A: Named Entity Recognition in RAFAEL}

As mentioned in section \ref{er}, apart from DeepER, RAFAEL employs also traditional NER-based solutions for entity recognition: {\em NERF}, {\em Liner2} and {\em Quant}. Each of them uses its own typology of named entities, which covers only a part of the types, enumerated in section \ref{analysis}. Table \ref{tab:NER} shows a correspondence between these types. As we can see, there are a few problems:
\begin{enumerate}
\item Many of NE types are not covered by neither {\em NERF} nor {\em Liner2},
\item For all geographical types, {\em NERF} has only one type {\em geogName}, which may affect QA precision,
\item In case of {\em CENTURY} and {\em YEAR}, {\em NERF} recognizes only a more general type {\em date}, from which they may be inferred,
\item {\em Liner2} does not differentiate between {\em NAME} and {\em SURNAME}, classifying both as parts of {\em person\_nam}.
\end{enumerate}
The problems 3 and 4 are solved by an additional postprocessing code, extracting {\em CENTURY} from {\em date} and {\em NAME} and {\em SURNAME} from {\em person\_nam} entities. In case of multi-segment person entities it assumes that the first and last word correspond to first and last name, respectively.
\begin{table}
\begin{tabular}{  r  c  c  c }
\hline
\hline
\textbf{Question NE type} & \textbf{NERF type} & \textbf{Liner2 type} & \textbf{Quant type}\\
\hline
PLACE & placeName:* & & \\ \hline
CONTINENT & \multirow{9}{*}{geogName}  &continent\_nam & \\
RIVER& &river\_nam & \\
LAKE& & & \\
MOUNTAIN& & & \\
RANGE& & mountain\_nam& \\
ISLAND& & island\_nam& \\
ARCHIPELAGO& & & \\
SEA& & sea\_nam& \\
CELESTIAL\_BODY& &astronomical\_nam & \\ \hline
COUNTRY&placeName:country &country\_nam & \\ \hline
\multirow{5}{*}{STATE}&\multirow{5}{*}{placeName:region} &admin1\_nam & \\ 
& &admin2\_nam & \\ 
& &admin3\_nam & \\
& &historical\_region\_nam & \\ 
& &country\_region\_nam & \\ \hline
CITY&placeName:settlement &city\_nam & \\ \hline
NATIONALITY&placeName:country & nation\_nam& \\ \hline
PERSON&persName &\multirow{3}{*}{person\_nam} & \\
NAME&persName:forename & &  \\
SURNAME&persName:surname & & \\ \hline
BAND&orgName &band\_nam & \\ \hline
DYNASTY&persName:addName & & \\ \hline
\multirow{3}{*}{ORGANISATION}&\multirow{3}{*}{orgName} &organization\_nam & \\
& &institution\_nam & \\ 
& &political\_party\_nam & \\ \hline
COMPANY&orgName &company\_nam & \\ \hline
EVENT&  & event\_nam& \\ \hline
TIME&\multirow{3}{*}{date}& &  \\
CENTURY&& &  \\
YEAR&& & \\ \hline
PERIOD&  & & quantity\\ \hline
COUNT&  & & number\\ \hline
QUANTITY&  & &quantity \\ \hline
VEHICLE&  & & \\ \hline
ANIMAL&  & & \\ \hline
\multirow{2}{*}{TITLE}&  &title\_nam& \\ 
&  &media\_nam & \\
\hline
\hline
\end{tabular}
  \caption{Correspondence between named entity types from question analysis and supported by different NER solutions.}
  \label{tab:NER}
\end{table}

While {\em NERF} and {\em Liner2} are standalone NER tools and details of their design are available in previously mentioned publications, {\em Quant} has been created specifically for RAFAEL. To find numbers, it annotates all chains of segments according to a predefined pattern, which accepts the following types of segments:
\begin{enumerate}
\item \textbf{(0-9)+} -- a string consisting of digits only,
\item \textbf{.} -- a period; a digit group separator in Polish ,
\item \textbf{,} -- a comma; a decimal mark in Polish ,
\item \textbf{num} -- a verbal expression of number, i.e. segments tagged as numerals.
\end{enumerate}
The pattern is matched in greedy mode, i.e. it adds as many new segments as possible. It could recognise expressions like \textit{10 tysięcy} (10 thousand), \textit{kilka milionów} (several million), \textit{10 000} or \textit{1.698,88} (1,698.88).

Quantity is a sequence of segments, recognised as a number, followed by a unit of measurement. To check whether a word denotes a unit of measurement, the {\em plWordNet} is searched for lexemes equal to its base. Then it suffices to check whether it belongs to a synset, having <jednostka miary 1> (unit of measurement) as one of (direct or indirect) hypernyms, e.g. \textit{piętnaście kilogramów} (fifteen kilograms) or \textit{5 000 watów} (5 000 watts).

\section*{Acknowledgments}
Study was supported by research fellowship within "Information technologies: research and their interdisciplinary applications" agreement number POKL.04.01.01-00-051/10-00. Critical reading of the manuscript by Agnieszka Mykowiecka and Aleksandra Brzezińska is gratefully acknowledged.

\bibliographystyle{newapa}
\bibliography{deeper.bib}

\begin{thebibliography}{}

\bibitem[\protect\citeauthoryear{Acedański}{Acedański}{2010}]{Acedanski2010}
Acedański, S. (2010).
\newblock {A morphosyntactic Brill Tagger for inflectional languages}.
\newblock In {\em Proceedings of the 7th international conference on Advances
  in Natural Language Processing (IceTAL'10 )}, (pp.\ 3--14). Springer-Verlag.

\bibitem[\protect\citeauthoryear{Ahn, Jijkoun, Mishne, M\"{u}ller, {De Rijke}
  \& Schlobach}{Ahn et~al.}{2004}]{Ahn2004}
Ahn, D., Jijkoun, V., Mishne, G., M\"{u}ller, K., {De Rijke}, M., \& Schlobach,
  S. (2004).
\newblock {Using Wikipedia at the TREC QA Track}.
\newblock In Voorhees, E.~M. \& Buckland, L.~P. (Eds.), {\em Proceedings of The
  Thirteenth Text REtrieval Conference (TREC 2004)}.

\bibitem[\protect\citeauthoryear{Armenska, Tomovski, Zdravkova \&
  Pehcevski}{Armenska et~al.}{2010}]{Armenska2010}
Armenska, J., Tomovski, A., Zdravkova, K., \& Pehcevski, J. (2010).
\newblock {Information Retrieval Using a Macedonian Test Collection for
  Question Answering}.
\newblock In {\em Proceedings of the 2nd International Conference ICT
  Innovations}, (pp.\ 205--214). Springer-Verlag.

\bibitem[\protect\citeauthoryear{Balasuriya, Ringland, Nothman, Murphy \&
  Curran}{Balasuriya et~al.}{2009}]{Balasuriya2009}
Balasuriya, D., Ringland, N., Nothman, J., Murphy, T., \& Curran, J.~R. (2009).
\newblock {Named entity recognition in Wikipedia}.
\newblock In {\em Proceedings of the 2009 Workshop on The People's Web Meets
  NLP: Collaboratively Constructed Semantic Resources}, (pp.\ 10--18).,
  Association for Computational Linguistics.

\bibitem[\protect\citeauthoryear{Brill, Dumais \& Banko}{Brill
  et~al.}{2002}]{Brill2002a}
Brill, E., Dumais, S., \& Banko, M. (2002).
\newblock {An analysis of the AskMSR question-answering system}.
\newblock In {\em Proceedings of the ACL-02 conference on Empirical methods in
  natural language processing - EMNLP '02}, volume~10, (pp.\ 257--264).
  Association for Computational Linguistics.

\bibitem[\protect\citeauthoryear{Ciaramita \& Altun}{Ciaramita \&
  Altun}{2006}]{Ciaramita2006}
Ciaramita, M. \& Altun, Y. (2006).
\newblock {Broad-coverage sense disambiguation and information extraction with
  a supersense sequence tagger}.
\newblock In {\em Proceedings of the 2006 Conference on Empirical Methods in
  Natural Language Processing - EMNLP '06}, (pp.\ 594). Association for
  Computational Linguistics.

\bibitem[\protect\citeauthoryear{Dakka \& Cucerzan}{Dakka \&
  Cucerzan}{2008}]{Dakka2008}
Dakka, W. \& Cucerzan, S. (2008).
\newblock {Augmenting Wikipedia with Named Entity Tags}.
\newblock In {\em Proceedings of the Third International Joint Conference on
  Natural Language Processing (IJCNLP 2008)}.

\bibitem[\protect\citeauthoryear{Dang, Kelly \& Lin}{Dang
  et~al.}{2007}]{Dang2008}
Dang, H.~T., Kelly, D., \& Lin, J. (2007).
\newblock {Overview of the TREC 2007 Question Answering track}.
\newblock In {\em Proceedings of The Sixteenth Text REtrieval Conference, TREC
  2007}.

\bibitem[\protect\citeauthoryear{Deg\'{o}rski}{Deg\'{o}rski}{2012}]{Degorski2012}
Deg\'{o}rski, L. (2012).
\newblock {Towards the Lemmatisation of Polish Nominal Syntactic Groups Using a
  Shallow Grammar}.
\newblock In Bouvry, P., Kłopotek, M.~A., Lepr\'{e}vost, F., Marciniak, M.,
  Mykowiecka, A., \& Rybiński, H. (Eds.), {\em Proceedings of the
  International Joint Conference on Security and Intelligent Information
  Systems}, volume 7053 of {\em Lecture Notes in Computer Science}, (pp.\
  370--378). Springer-Verlag.

\bibitem[\protect\citeauthoryear{Duclaye, Sitko, Filoche \& Collin}{Duclaye
  et~al.}{2002}]{Duclaye2002}
Duclaye, F., Sitko, J., Filoche, P., \& Collin, O. (2002).
\newblock {A Polish Question-Answering System for Business Information}.
\newblock In {\em BIS 2002, 5th International Conference on Business
  Information Systems, Poznań, Poland, 24-25 April 2002}, (pp.\ 209--212).

\bibitem[\protect\citeauthoryear{Ferrucci, Brown, Chu-carroll, Fan, Gondek,
  Kalyanpur, Lally, Murdock, Nyberg, Prager, Schlaefer \& Welty}{Ferrucci
  et~al.}{2010}]{Ferrucci2010}
Ferrucci, D.~A., Brown, E., Chu-carroll, J., Fan, J., Gondek, D., Kalyanpur,
  A.~A., Lally, A., Murdock, J.~W., Nyberg, E., Prager, J., Schlaefer, N., \&
  Welty, C. (2010).
\newblock {Building Watson: An Overview of the DeepQA Project}.
\newblock {\em AI Magazine}, {\em 31\/}(3), 59--79.

\bibitem[\protect\citeauthoryear{Galambos}{Galambos}{2001}]{Galambos2001}
Galambos, L. (2001).
\newblock {Lemmatizer for Document Information Retrieval Systems in JAVA}.
\newblock In {\em Proceedings of the 28th Conference on Current Trends in
  Theory and Practice of Informatics (SOFSEM 2001)}, (pp.\ 243--252).
  Springer-Verlag.

\bibitem[\protect\citeauthoryear{Harabagiu, Moldovan, Pasca, Mihalcea,
  Surdeanu, Bunescu, G\^{\i}rju, Rus \& Morarescu}{Harabagiu
  et~al.}{2001}]{Harabagiu2001}
Harabagiu, S., Moldovan, D., Pasca, M., Mihalcea, R., Surdeanu, M., Bunescu,
  R., G\^{\i}rju, R., Rus, V., \& Morarescu, P. (2001).
\newblock {The role of lexico-semantic feedback in open-domain textual
  question-answering}.
\newblock In {\em Proceedings of the 39th Annual Meeting on Association for
  Computational Linguistics - ACL '01}, (pp.\ 282--289). Association for
  Computational Linguistics.

\bibitem[\protect\citeauthoryear{Hovy, Gerber, Hermjakob, Junk \& Lin}{Hovy
  et~al.}{2000}]{Hovy2000}
Hovy, E., Gerber, L., Hermjakob, U., Junk, M., \& Lin, C.-Y. (2000).
\newblock {Question Answering in Webclopedia}.
\newblock In {\em Proceedings of The Ninth Text REtrieval Conference (TREC
  2000)}.

\bibitem[\protect\citeauthoryear{Jaccard}{Jaccard}{1901}]{Jaccard1901}
Jaccard, P. (1901).
\newblock {\'{E}tude comparative de la distribution florale dans une portion
  des Alpes et des Jura}.
\newblock {\em Bulletin del la Soci\'{e}t\'{e} Vaudoise des Sciences
  Naturelles}, {\em 37}, 547--579.

\bibitem[\protect\citeauthoryear{Karzewski}{Karzewski}{1997}]{Karzewski1997}
Karzewski, M. (1997).
\newblock {\em {Jeden z dziesięciu - pytania i odpowiedzi}}.
\newblock Muza SA.

\bibitem[\protect\citeauthoryear{Katz, Lin, Loreto, Hildebrandt, Bilotti,
  Felshin, Fernandes, Marton \& Mora}{Katz et~al.}{2003}]{Katz2003}
Katz, B., Lin, J., Loreto, D., Hildebrandt, W., Bilotti, M., Felshin, S.,
  Fernandes, A., Marton, G., \& Mora, F. (2003).
\newblock {Integrating Web-based and corpus-based techniques for question
  answering}.
\newblock In {\em Proceedings of the Twelfth Text REtrieval Conference (TREC
  2003)}.

\bibitem[\protect\citeauthoryear{Kazama \& Torisawa}{Kazama \&
  Torisawa}{2007}]{Kazama2007}
Kazama, J. \& Torisawa, K. (2007).
\newblock {Exploiting Wikipedia as External Knowledge for Named Entity
  Recognition}.
\newblock In {\em In Joint Conference on Empirical Methods in Natural Language
  Processing and Computational Natural Language Learning (2007)}, (pp.\
  698--707). Association for Computational Linguistics.

\bibitem[\protect\citeauthoryear{Konop\'{\i}k \& Rohl\'{\i}k}{Konop\'{\i}k \&
  Rohl\'{\i}k}{2010}]{Konopik2010}
Konop\'{\i}k, M. \& Rohl\'{\i}k, O. (2010).
\newblock {Question Answering for Not Yet Semantic Web}.
\newblock In {\em Proceedings of the 13th International Conference on Text,
  Speech and Dialogue (TSD 2010)}, (pp.\ 125--132). Springer-Verlag.

\bibitem[\protect\citeauthoryear{Kope\'{c} \& Ogrodniczuk}{Kope\'{c} \&
  Ogrodniczuk}{2012}]{Kopec2012}
Kope\'{c}, M. \& Ogrodniczuk, M. (2012).
\newblock {Creating a Coreference Resolution System for Polish}.
\newblock In {\em The eighth international conference on Language Resources and
  Evaluation (LREC)}. European Language Resources Association (ELRA).

\bibitem[\protect\citeauthoryear{Lee, Wang, Kim \& Jang}{Lee
  et~al.}{2005}]{Lee2005}
Lee, C., Wang, J.-H., Kim, H.-J., \& Jang, M.-G. (2005).
\newblock {Extracting Template for Knowledge-based Question-Answering Using
  Conditional Random Fields}.
\newblock In {\em Proceedings of the 28th Annual International ACM SIGIR
  Workshop on MFIR}, (pp.\ 428--434). ACM.

\bibitem[\protect\citeauthoryear{Li \& Roth}{Li \& Roth}{2002}]{Li2002}
Li, X. \& Roth, D. (2002).
\newblock {Learning Question Classifiers}.
\newblock In {\em Proceedings of the 19th International Conference on
  Computational Linguistics (COLING-2002)}, volume~1 of {\em COLING '02}.
  Association for Computational Linguistics.

\bibitem[\protect\citeauthoryear{Lombarovi\'{c}, \v{S}najder \&
  Ba\v{s}i\'{c}}{Lombarovi\'{c} et~al.}{2011}]{Lombarovic2011}
Lombarovi\'{c}, T., \v{S}najder, J., \& Ba\v{s}i\'{c}, B.~D. (2011).
\newblock {Question Classification for a Croatian QA System}.
\newblock In {\em Proceedings of the 14th International Conference on Text,
  Speech and Dialogue (TSD 2011)}, (pp.\ 403--410). Springer-Verlag.

\bibitem[\protect\citeauthoryear{Mann}{Mann}{2002}]{Mann2002}
Mann, G.~S. (2002).
\newblock {Fine-grained proper noun ontologies for question answering}.
\newblock In {\em Proceedings of the 2002 workshop on Building and using
  semantic networks (SEMANET '02)}, volume~11, (pp.\ 1--7). Association for
  Computational Linguistics.

\bibitem[\protect\citeauthoryear{Marcińczuk \& Janicki}{Marcińczuk \&
  Janicki}{2012}]{Marcinczuk2012}
Marcińczuk, M. \& Janicki, M. (2012).
\newblock {Optimizing CRF-based Model for Proper Name Recognition in Polish
  Texts}.
\newblock In {\em Proceedings of CICLing 2012, Part I}, (pp.\ 258--269).
  Springer-Verlag.

\bibitem[\protect\citeauthoryear{Marcińczuk, Ptak, Radziszewski \&
  Piasecki}{Marcińczuk et~al.}{2013}]{Marcinczuk2013}
Marcińczuk, M., Ptak, M., Radziszewski, A., \& Piasecki, M. (2013).
\newblock {Open dataset for development of Polish Question Answering systems}.
\newblock In {\em Proceedings of the 6th Language \& Technology Conference:
  Human Language Technologies as a Challenge for Computer Science and
  Linguistics}. Wydawnictwo Poznańskie, Fundacja Uniwersytetu im. Adama
  Mickiewicza.

\bibitem[\protect\citeauthoryear{Marcińczuk, Radziszewski, Piasecki, Piasecki
  \& Ptak}{Marcińczuk et~al.}{2013}]{Radziszewski2013}
Marcińczuk, M., Radziszewski, A., Piasecki, M., Piasecki, D., \& Ptak, M.
  (2013).
\newblock {Evaluation of a Baseline Information Retrieval for a Polish
  Open-domain Question Answering System}.
\newblock In {\em Proceedings of the International Conference Recent Advances
  in Natural Language Processing (RANLP 2013)}, (pp.\ 428--435). Association
  for Computational Linguistics.

\bibitem[\protect\citeauthoryear{Maziarz, Piasecki \& Szpakowicz}{Maziarz
  et~al.}{2012}]{Maziarz2012}
Maziarz, M., Piasecki, M., \& Szpakowicz, S. (2012).
\newblock {Approaching plWordNet 2.0}.
\newblock In {\em Proceedings of the 6th Global Wordnet Conference}.

\bibitem[\protect\citeauthoryear{Moldovan, Harabagiu, Paşca, Mihalcea,
  G\^{\i}rju, Goodrum \& Rus}{Moldovan et~al.}{2000}]{Moldovan2000}
Moldovan, D., Harabagiu, S., Paşca, M., Mihalcea, R., G\^{\i}rju, R., Goodrum,
  R., \& Rus, V. (2000).
\newblock {The structure and performance of an open-domain question answering
  system}.
\newblock In {\em Proceedings of the 38th Annual Meeting on Association for
  Computational Linguistics - ACL '00}, (pp.\ 563--570). Association for
  Computational Linguistics.

\bibitem[\protect\citeauthoryear{Moldovan \& Novischi}{Moldovan \&
  Novischi}{2002}]{Moldovan2002a}
Moldovan, D. \& Novischi, A. (2002).
\newblock {Lexical chains for question answering}.
\newblock In {\em Proceedings of the 19th International Conference on
  Computational Linguistics (COLING-2002)}. Association for Computational
  Linguistics.

\bibitem[\protect\citeauthoryear{Oh, Torisawa, Hashimoto, Sano, Saeger \&
  Ohtake}{Oh et~al.}{2013}]{Oh2013}
Oh, J.-h., Torisawa, K., Hashimoto, C., Sano, M., Saeger, S.~D., \& Ohtake, K.
  (2013).
\newblock {Why-Question Answering using Intra- and Inter-Sentential Causal
  Relations}.
\newblock In {\em Proceedings of the 51st Annual Meeting of the Association for
  Computational Linguistics}, (pp.\ 1733--1743). Association for Computational
  Linguistics.

\bibitem[\protect\citeauthoryear{Osenova, Simov, Simov, Tanev \&
  Kouylekov}{Osenova et~al.}{2004}]{Osenova2004}
Osenova, P., Simov, A., Simov, K., Tanev, H., \& Kouylekov, M. (2004).
\newblock {Bulgarian-english question answering: adaptation of language
  resources}.
\newblock In Peters, C., Clough, P., Gonzalo, J., Jones, G. J.~F., Kluck, M.,
  \& Magnini, B. (Eds.), {\em Proceedings of the 5th conference on
  Cross-Language Evaluation Forum: multilingual Information Access for Text,
  Speech and Images (CLEF'04)}, volume 3491 of {\em Lecture Notes in Computer
  Science}, (pp.\ 458--469). Springer-Verlag.

\bibitem[\protect\citeauthoryear{Peshterliev \& Koychev}{Peshterliev \&
  Koychev}{2011}]{Peshterliev2011}
Peshterliev, S. \& Koychev, I. (2011).
\newblock {Semantic Retrieval Approach to Factoid Question Answering for
  Bulgarian}.
\newblock In {\em Proceedings of the 3rd International Conference on Software,
  Services and Semantic Technologies (S3T 2011)}, (pp.\ 25--32).
  Springer-Verlag.

\bibitem[\protect\citeauthoryear{Piechociński \& Mykowiecka}{Piechociński \&
  Mykowiecka}{2005}]{Piechocinski2005}
Piechociński, D. \& Mykowiecka, A. (2005).
\newblock {Question answering in Polish using shallow parsing}.
\newblock In Garab\'{\i}k, R. (Ed.), {\em Computer Treatment of Slavic and East
  European Languages: Proceedings of the Third International Seminar,
  Bratislava, Slovakia}, (pp.\ 167--173). VEDA: Vydava- tel’stvo Slovenskej
  akad\'{e}me vied.

\bibitem[\protect\citeauthoryear{Ponzetto \& Strube}{Ponzetto \&
  Strube}{2007}]{Ponzetto2007}
Ponzetto, S.~P. \& Strube, M. (2007).
\newblock {Deriving a Large Scale Taxonomy from Wikipedia}.
\newblock {\em Artificial Intelligence}, {\em 22}, 1440--1445.

\bibitem[\protect\citeauthoryear{Przepi\'{o}rkowski}{Przepi\'{o}rkowski}{2007}]{Przepiorkowski2007}
Przepi\'{o}rkowski, A. (2007).
\newblock {Slavonic information extraction and partial parsing}.
\newblock In {\em Proceedings of the Workshop on Balto-Slavonic Natural
  Language Processing Information Extraction and Enabling Technologies - ACL
  '07}. Association for Computational Linguistics.

\bibitem[\protect\citeauthoryear{Przepi\'{o}rkowski}{Przepi\'{o}rkowski}{2008}]{Przepiorkowski2008}
Przepi\'{o}rkowski, A. (2008).
\newblock {\em {Powierzchniowe przetwarzanie języka polskiego}}.
\newblock Warszawa: Akademicka Oficyna Wydawnicza EXIT.

\bibitem[\protect\citeauthoryear{Przepi\'{o}rkowski, Bańko, G\'{o}rski \&
  Lewandowska-Tomaszczyk}{Przepi\'{o}rkowski et~al.}{2012}]{Przepiorkowski2012}
Przepi\'{o}rkowski, A., Bańko, M., G\'{o}rski, R.~L., \&
  Lewandowska-Tomaszczyk, B. (2012).
\newblock {\em {Narodowy Korpus Języka Polskiego}}.
\newblock Warszawa: Wydawnictwo Naukowe PWN.

\bibitem[\protect\citeauthoryear{Przybyła}{Przybyła}{2012}]{Przybya2012}
Przybyła, P. (2012).
\newblock {Issues of Polish Question Answering}.
\newblock In Hryniewicz, O., Mielniczuk, J., Penczek, W., \& Waniewski, J.
  (Eds.), {\em Proceedings of the first conference 'Information Technologies:
  Research and their Interdisciplinary Applications' (ITRIA 2012)}, (pp.\
  122--139)., Warsaw. Institute of Computer Science, Polish Academy of
  Sciences.

\bibitem[\protect\citeauthoryear{Przybyła}{Przybyła}{2013a}]{Przybya2013}
Przybyła, P. (2013a).
\newblock {Question Analysis for Polish Question Answering}.
\newblock In {\em 51st Annual Meeting of the Association for Computational
  Linguistics, Proceedings of the Student Research Workshop}, (pp.\ 96--102).,
  Sofia, Bulgaria. Association for Computational Linguistics.

\bibitem[\protect\citeauthoryear{Przybyła}{Przybyła}{2013b}]{Przybyla}
Przybyła, P. (2013b).
\newblock {Question Classification for Polish Question Answering}.
\newblock In Kłopotek, M.~A., Koronacki, J., Marciniak, M., Mykowiecka, A., \&
  Wierzchoń, S.~T. (Eds.), {\em Proceedings of the 20th International
  Conference on Language Processing and Intelligent Information Systems
  (LP\&IIS 2013)}, (pp.\ 50--56). Springer-Verlag.

\bibitem[\protect\citeauthoryear{Richman \& Schone}{Richman \&
  Schone}{2008}]{Richman2008}
Richman, A.~E. \& Schone, P. (2008).
\newblock {Mining Wiki Resources for Multilingual Named Entity Recognition}.
\newblock In {\em Proceedings of the 46th Annual Meeting of the Association for
  Computational Linguistics (ACL 2008)}. Association for Computational
  Linguistics.

\bibitem[\protect\citeauthoryear{Ruiz-Casado, Alfonseca \&
  Castells}{Ruiz-Casado et~al.}{2005}]{Ruiz-Casado2005}
Ruiz-Casado, M., Alfonseca, E., \& Castells, P. (2005).
\newblock {Automatic Assignment of Wikipedia Encyclopedic Entries to WordNet
  Synsets}.
\newblock {\em Advances in Web Intelligence}, {\em 3528}, 380--386.

\bibitem[\protect\citeauthoryear{Savary, Manicki \& Baron}{Savary
  et~al.}{2013}]{Savary2013}
Savary, A., Manicki, L., \& Baron, M. (2013).
\newblock {Populating a multilingual ontology of proper names from open
  sources}.
\newblock {\em Journal of Language Modelling}, {\em 1\/}(2), 189--225.

\bibitem[\protect\citeauthoryear{Savary \& Waszczuk}{Savary \&
  Waszczuk}{2012}]{Savary2012}
Savary, A. \& Waszczuk, J. (2012).
\newblock {Narzędzia do anotacji jednostek nazewniczych}.
\newblock In {\em Narodowy Korpus Języka Polskiego [Eng.: National Corpus of
  Polish]}  (pp.\ 225--252). Wydawnictwo Naukowe PWN.

\bibitem[\protect\citeauthoryear{Shapiro}{Shapiro}{1992}]{Shapiro1992}
Shapiro, S.~C. (1992).
\newblock {\em {Encyclopedia of Artificial Intelligence}}.
\newblock John Wiley \& Sons, Inc.

\bibitem[\protect\citeauthoryear{Simov \& Osenova}{Simov \&
  Osenova}{2005}]{Simov2005}
Simov, K. \& Osenova, P. (2005).
\newblock {BulQA: Bulgarian–bulgarian question answering at CLEF 2005}.
\newblock In Peters, C., Gey, F.~C., Gonzalo, J., M\"{u}ller, H., Jones, G.
  J.~F., Kluck, M., Magnini, B., \& Rijke, M. (Eds.), {\em Proceedings of the
  6th international conference on Cross-Language Evalution Forum: accessing
  Multilingual Information Repositories (CLEF'05)}, volume 4022 of {\em Lecture
  Notes in Computer Science}, (pp.\ 517--526). Springer-Verlag.

\bibitem[\protect\citeauthoryear{Solovyev}{Solovyev}{2013}]{Solovyev2013}
Solovyev, A. (2013).
\newblock {Dependency-Based Algorithms for Answer Validation Task in Russian
  Question Answering}.
\newblock In {\em Proceedings of the 25th International Conference on Language
  Processing and Knowledge in the Web (GSCL 2013)}, (pp.\ 199--212).
  Springer-Verlag.

\bibitem[\protect\citeauthoryear{Tanev}{Tanev}{2004}]{Tanev2004}
Tanev, H. (2004).
\newblock {Socrates - a Question Answering prototype for Bulgarian}.
\newblock In N.~Nikolov (Ed.), {\em Recent Advances in Natural Language
  Processing: Selected Papers from RANLP 2003}  (pp.\ 377--386). John
  Benjamins.

\bibitem[\protect\citeauthoryear{Toral \& Mu\~{n}oz}{Toral \&
  Mu\~{n}oz}{2006}]{Toral2006}
Toral, A. \& Mu\~{n}oz, R. (2006).
\newblock {A proposal to automatically build and maintain gazetteers for Named
  Entity Recognition by using Wikipedia}.
\newblock In {\em Proceedings of the 11th Conference of the European Chapter of
  the Association for Computational Linguistics}. Association for Computational
  Linguistics.

\bibitem[\protect\citeauthoryear{Toral, Mu\~{n}oz \& Monachini}{Toral
  et~al.}{2008}]{Toral2008}
Toral, A., Mu\~{n}oz, R., \& Monachini, M. (2008).
\newblock {Named Entity WordNet}.
\newblock In {\em Proceedings of the International Conference on Language
  Resources and Evaluation, LREC 2008}.

\bibitem[\protect\citeauthoryear{\v{C}eh \& Ojster\v{s}ek}{\v{C}eh \&
  Ojster\v{s}ek}{2009}]{Ceh2009}
\v{C}eh, I. \& Ojster\v{s}ek, M. (2009).
\newblock {Developing a question answering system for the slovene language}.
\newblock {\em WSEAS Transactions on Information Science and Applications},
  {\em 6\/}(9), 1533--1543.

\bibitem[\protect\citeauthoryear{Vetulani}{Vetulani}{1988}]{Vetulani1988}
Vetulani, Z. (1988).
\newblock {PROLOG Implementation of an Access in Polish to a Data Base}.
\newblock In {\em Studia z automatyki, XII}  (pp.\ 5--23). PWN.

\bibitem[\protect\citeauthoryear{Walas}{Walas}{2012}]{Walas2012}
Walas, M. (2012).
\newblock {How to answer yes/no spatial questions using qualitative reasoning?}
\newblock In Gelbukh, A. (Ed.), {\em 13th International Conference on
  Computational Linguistics and Intelligent Text Processing}, (pp.\ 330--341).
  Springer-Verlag.

\bibitem[\protect\citeauthoryear{Walas \& Jassem}{Walas \&
  Jassem}{2010}]{Walas2010}
Walas, M. \& Jassem, K. (2010).
\newblock {Named entity recognition in a Polish question answering system}.
\newblock In {\em Intelligent Information Systems}, (pp.\ 181--191). Publishing
  House of University of Podlasie.

\bibitem[\protect\citeauthoryear{Walas \& Jassem}{Walas \&
  Jassem}{2011}]{Walas2011}
Walas, M. \& Jassem, K. (2011).
\newblock {Spatial reasoning and disambiguation in the process of knowledge
  acquisition}.
\newblock In {\em Proceedings of the 5th Language \& Technology Conference:
  Human Language Technologies as a Challenge for Computer Science and
  Linguistics}, (pp.\ 420--424). Fundacja Uniwersytetu im. Adama Mickiewicza.

\bibitem[\protect\citeauthoryear{Woliński}{Woliński}{2006}]{Wolinski2006a}
Woliński, M. (2006).
\newblock {Morfeusz — a Practical Tool for the Morphological Analysis of
  Polish}.
\newblock In M.~Kłopotek, S.~Wierzchoń, \& K.~Trojanowski (Eds.), {\em
  Intelligent Information Processing and Web Mining}  (pp.\ 511--520).
  Springer-Verlag.

\bibitem[\protect\citeauthoryear{Yih, Chang, Meek \& Pastusiak}{Yih
  et~al.}{2013}]{Yih2013}
Yih, W.-t., Chang, M.-w., Meek, C., \& Pastusiak, A. (2013).
\newblock {Question Answering Using Enhanced Lexical Semantic Models}.
\newblock In {\em Proceedings of the 51st Annual Meeting of the Association for
  Computational Linguistics}, (pp.\ 1744--1753). Association for Computational
  Linguistics.

\end{thebibliography}

\end{document}